\documentclass[lettersize,journal]{IEEEtran}      

\usepackage{amsmath,amsfonts}
\usepackage{algorithmic}
\usepackage{algorithm}
\usepackage{array}
\usepackage[caption=false,font=normalsize,labelfont=sf,textfont=sf]{subfig}
\usepackage{textcomp}
\usepackage{stfloats}
\usepackage{url}
\usepackage{verbatim}
\usepackage{graphicx}
\usepackage{cite}
\usepackage{todonotes}
\usepackage{float}
\usepackage{multirow}
\usepackage{mdframed}
\usepackage{xfrac}
\usepackage{orcidlink}
\usepackage{hyperref}

\newcommand{\addedText}[1]{\textcolor{black}{#1}} 

\begin{document}
\title{
    Biomechanics-Aware Trajectory Optimization for Online Navigation during Robotic Physiotherapy}

\author{Italo Belli \orcidlink{0000-0002-8064-2251}\textsuperscript{1,2*}, Florian van Melis\textsuperscript{2}, J. Micah Prendergast \orcidlink{0000-0002-9888-3133}\textsuperscript{1}, Ajay Seth \orcidlink{0000-0003-4217-1580}\textsuperscript{2}, and Luka Peternel \orcidlink{0000-0002-8696-3689}\textsuperscript{1}
\thanks{The authors are with the \textsuperscript{1}Cognitive Robotics and the \textsuperscript{2}BioMechanical Engineering departments, Delft University of Technology, Delft, The Netherlands (email: i.belli@tudelft.nl, f.j.vanmelis@student.tudelft.nl, j.m.prendergast@tudelft.nl, a.seth@tudelft.nl, l.peternel@tudelft.nl)}
\thanks{\textsuperscript{*}Corresponding Author: Italo Belli}
}



\maketitle

\begin{abstract}

Robotic devices provide a great opportunity to assist in delivering physical therapy and rehabilitation movements, yet current robot-assisted methods struggle to incorporate biomechanical metrics essential for safe and effective therapy. 
We introduce BATON, a Biomechanics-Aware Trajectory Optimization approach to online robotic Navigation of human musculoskeletal loads for rotator cuff rehabilitation. BATON embeds a high-fidelity OpenSim model of the human shoulder into an optimal control framework, generating strain-minimizing trajectories for real-time control of therapeutic movements. \addedText{Its core strength lies in the ability to adapt biomechanics-informed trajectories online to unpredictable volitional human actions or reflexive reactions during physical human-robot interaction based on robot-sensed motion and forces. BATON's adaptability is enabled by a real-time, model-based estimator that infers changes in muscle activity via a rapid redundancy solver driven by robot pose and force/torque sensor data. We validated BATON through physical human-robot interaction experiments, assessing response speed, motion smoothness, and interaction forces.}

\end{abstract}



\section{Introduction}

\IEEEPARstart{I}{njuries} affecting the musculoskeletal system are pervasive in society due to an increasingly aging population, strenuous manual labor, and more widespread engagement in sports~\cite{minagawa2013prevalence, leclerc2004incidence, seroyer2009shoulder}. Disorders impacting the shoulder and the rotator cuff are particularly common, with clinical literature reporting their prevalence to be as high as $22\%$ in the general population~\cite{minagawa2013prevalence} and 50\% over age 66~\cite{codding2018natural}. Overall, the demand for therapy and rehabilitation is large and expected to grow, exacerbating the problem posed by a lack of physical therapists~\cite{zimbelman2010physical}. Moreover, the task of treating a complex biological mechanism, such as the shoulder, is challenging due to a fundamental lack of insights into rotator cuff behavior during rehabilitation. Physiotherapists need to move the joint safely, gradually increasing the patient's range of motion while physically supporting the patient's arm and avoiding re-injuries~\cite{thomson2016rehabilitation}. Consequently, simple and limited movements are prescribed, to limit both the risk of re-injury and the physical demands on the therapist to manually manipulate the patient safely. \addedText{However, overly conservative movements may result in a limited range of motion during the therapy, which can delay recovery since moving through
a larger range of motion was shown to expedite recovery~\cite{osteraas2013dose,proietti2016upper}}

Robotic technologies can assist in addressing manual manipulation and re-injury risks during therapy: robots can reduce the manual burden on physiotherapists and provide new tools to monitor and improve rehabilitation outcomes. Overall, rehabilitative robots are already used successfully in post-stroke rehabilitation~\cite{sommerhalder2023polymorphic, marchal2009review, calafiore2022efficacy} and gait assistance~\cite{nam2017robot, jamwal2020robotic, poggensee2021adaptation}. However, robots to treat musculoskeletal injuries (such as rotator cuff tears) remain limited. In particular, previous work remains focused on relieving physiotherapists from physically manipulating the patient. Smooth rehabilitation trajectories can be guaranteed by focusing on the control of the movement of the rehabilitative robot itself~\cite{cen2022trajectory}, and therapeutic robotic movements can be learned from expert human demonstrations to automate manual patient manipulation~\cite{luciani2023trajectory}. However, these solutions do not directly account for the patient's biomechanics. Deeper insights into human musculoskeletal mechanics can enable robots to be aware of tissue and joint loads associated with injury risks, allowing them to operate autonomously or in collaboration with therapists to holistically improve the rehabilitation process~\cite{survey_2023}. An increasingly promising strategy is to leverage human musculoskeletal models and incorporate them directly into robot control loops~\cite{carmichael2013estimating, ghannadi2017nonlinear, zignoli2019including, fang2020human, gordon2022human, li2022natural, gallois2025effort}.

\begin{figure}[t!]
\centering
\includegraphics[width=\columnwidth]{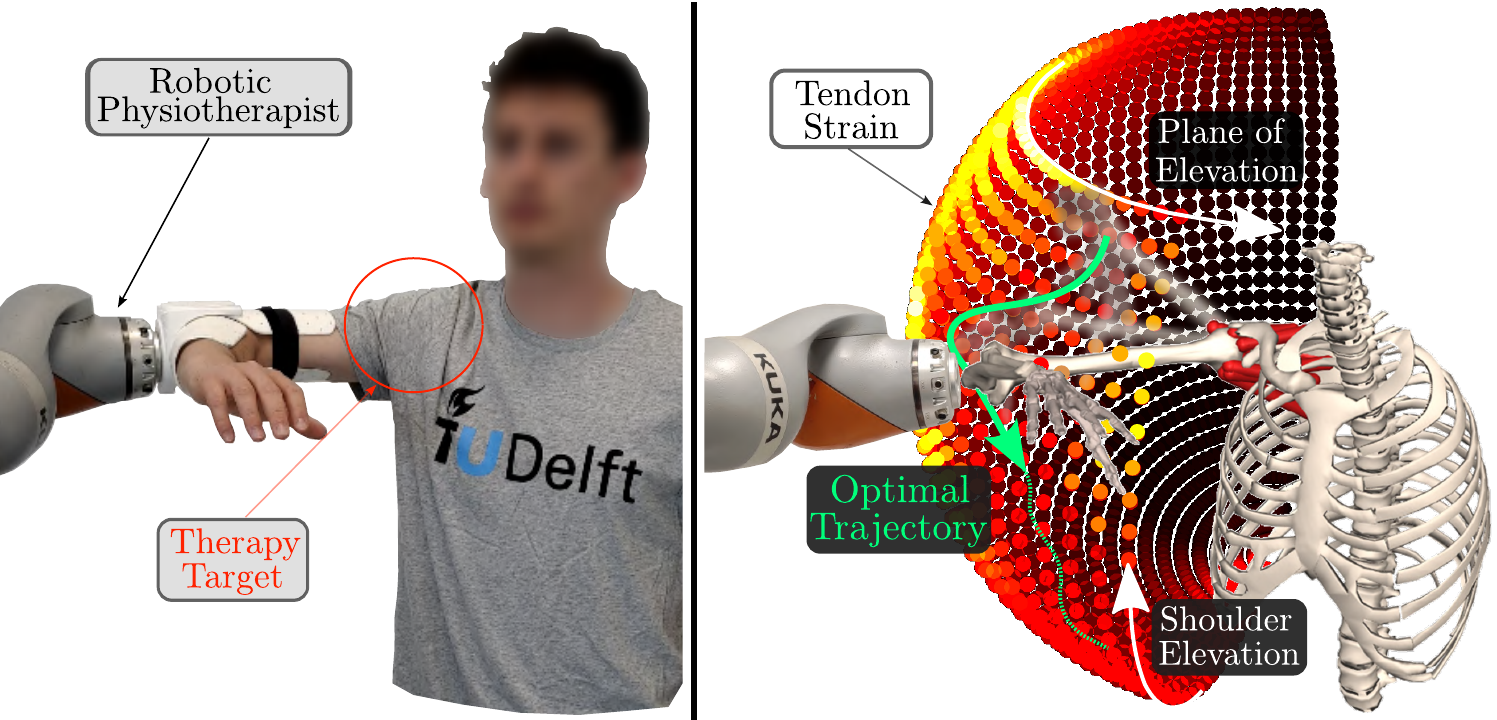}
\caption{BATON combines musculoskeletal modeling and human-robot interaction to enable a robotic physiotherapist to plan therapeutic movements for its patients in real-time while considering tissue loading (i.e. strain) induced in the rotator cuff tendons in the shoulder.}
\label{nicefigure}
\vspace{-0.4cm}
\end{figure}

Musculoskeletal modeling has progressed greatly over the past decades, allowing researchers to estimate the activities of individual muscles involved in producing movement~\cite{thelen2003generating, de2021perspective, belli2023does} and to perform predictive simulations of human motion~\cite{falisse2019rapid, dembia2020moco}. As computational models increasingly capture the inner workings of the human body, their applicability continues to expand, particularly in physical human-robot interaction and assistive devices. Recently, biomechanical models have been used to quantify assistance needed by a human operator~\cite{carmichael2013estimating, ghannadi2017nonlinear, zignoli2019including}, reduce human metabolic cost in walking-assistive devices~\cite{fang2020human, gordon2022human, li2022natural}, and even plan search-and-rescue robotic operations~\cite{peiros2023finding} or the motion of supernumerary robotics limbs~\cite{supernumerary2021}. 
Other researchers designed optimized trajectories for controlling an ankle rehabilitation robot, to minimize joint loading~\cite{jamwal2020musculoskeletal}. A custom ankle model was used to plan the robot's trajectory offline, making online model-free adjustments based on tracking errors attributed to excessive joint stiffness. 
Through computer simulations or offline usage of human models, these works demonstrated the importance and utility of including human biomechanics in robot planning and decision-making. However, they did not leverage high-fidelity biomechanical simulations to regulate human-robot interactions in scenarios that require online adaptation to human behavior, such as robotic physiotherapy.

When real-time capabilities are necessary, purely kinematic models have been used to monitor human joint positions, velocities, and torques~\cite{latella2021analysis}, or to develop controllers for hybrid neuroprostheses~\cite{bao2020tube}. While lower-fidelity models allow for computational efficiency, they are unsuitable for deep insights into the musculoskeletal system during physical human-robot interaction. 

Recently, biomechanical quantities from a high-fidelity human model have been used in an online model predictive control framework, achieving predictive control of a leg prosthesis for ergonomically safe walking~\cite{clark2022learning}. Researchers leveraged fast predictions of how knee torque relates to ankle angle, by including offline information from a musculoskeletal model in an imitation learning strategy. \addedText{Moya-Esteban and Sartori, instead, directly leveraged surface EMG-driven human models to minimize lumbar joint compression during external load handling, modulating the assistance of a wearable back exoskeleton online~\cite{moya2023real, sartori2025ceinms}.}

However, these studies did not consider the mechanical behavior of the human tissues (e.g., muscles and tendons) during physical human-robot interaction \addedText{for shoulder injuries. Without such consideration, safety cannot be ensured throughout the rehabilitation movements. Furthermore, surface EMG is limited to the measurement of surface muscles, while the underlying muscles and tissues cannot be monitored. Since typical shoulder injuries involve the rotator cuff muscles, which are deep-lying muscles that hold the arm in place, there is a need for insights into the underlying biological tissues.}

\addedText{To gain such underlying insights, tissue loads can be estimated from external measurements based on a high-fidelity model of the human shoulder. In this direction, researchers developed the concept of ``strain maps"~\cite{prendergast2021biomechanics} to intuitively represent musculoskeletal tissue loadings. A strain map} captures the relationship between human pose and the strain (i.e., load) induced in the rotator cuff tendons. By abstracting the rotator cuff tendon strains into an intuitive and navigable map, offline graph-based motion planning for safe robotic-mediated shoulder rehabilitation was achieved. However, the planning algorithm did not account for the dynamics of the human, resulting in trajectories that presented sudden direction changes, potentially difficult to track, and dynamically unsafe. 

\addedText{Moreover, the method in~\cite{prendergast2021biomechanics} did not consider online variations in tendon strain level caused by muscle activations during therapy, limiting its real-world applicability.}
Employing the strain maps, a reactive impedance-control-based approach to physical human-robot interaction in rehabilitation was also presented, to enable a subject to perform therapeutic exercises while robotic assistance protected them from navigating through potentially dangerous poses~\cite{balvert2022enabling}. In this case, the system would react only when dangerous movements were already initiated, without anticipating the subject's trajectory, which could lead to abrupt movements and high corrective forces from the robot. \addedText{Critically, unpredictable volitional actions of the patient and/or reflexive reactions to physical human-robot interaction in terms of muscle activity and their effects on tendon strains were not considered.}

\addedText{It is clear that musculoskeletal models have the potential to inform both the planning of the therapeutic motion, in regard to tissue loading, and the human activity and response to robot movement through physical human-robot interaction. However, there has been no single framework for combining planning, execution, and human response during robotic rehabilitation in an online manner.}

\addedText{To address this gap, our overarching} objective is to develop a Biomechanics-Aware Trajectory Optimization for the Navigation of \addedText{real-time} human tissue strains during robotic physiotherapy (BATON) targeting musculoskeletal rehabilitation. Our novel approach directly couples a state-of-the-art biomechanical model of the human shoulder to robotic control, generating movements for predictive robotic-assisted musculoskeletal rehabilitation in real-time. \addedText{This is achieved by accounting for changes in muscle activity due to unpredictable volitional actions of the patient and/or reflexive reaction to physical human-robot interaction that is estimated directly from robot sensors}. The resulting planned trajectories are realized by a collaborative robot arm that delivers the movement to a healthy human subject through safe impedance control \addedText{and provides estimations of contact forces and human poses to close the loop. We propose a muscle activity estimation that provides an alternative to sensor-based muscle activity estimation from surface EMG. Our system can inform the online path planning about the activity of deep-lying muscles, such as rotator cuffs, which cannot be measured with surface EMG.} 

\addedText{To achieve our objective, we target the} following specific aims:
\begin{itemize}

    \item A1: \addedText{design a biomechanics-aware trajectory optimization planner that efficiently incorporates a real-time shoulder musculoskeletal model into the underlying optimal control problem (OCP) by decoupling skeletal dynamics and tendon strain behavior of the rotator cuff muscles};



    \item A2: \addedText{develop a method to inform the planner of the dynamic changes in human biomechanics dependent on muscle activity during real-time physical human-robot interaction;}

    \item A3: \addedText{experimentally evaluate the performance of BATON in terms of the speed, smoothness, and forces of human-in-the-loop response.}
    
    
\end{itemize}


\begin{figure*}[t]
    \includegraphics[width=\textwidth]{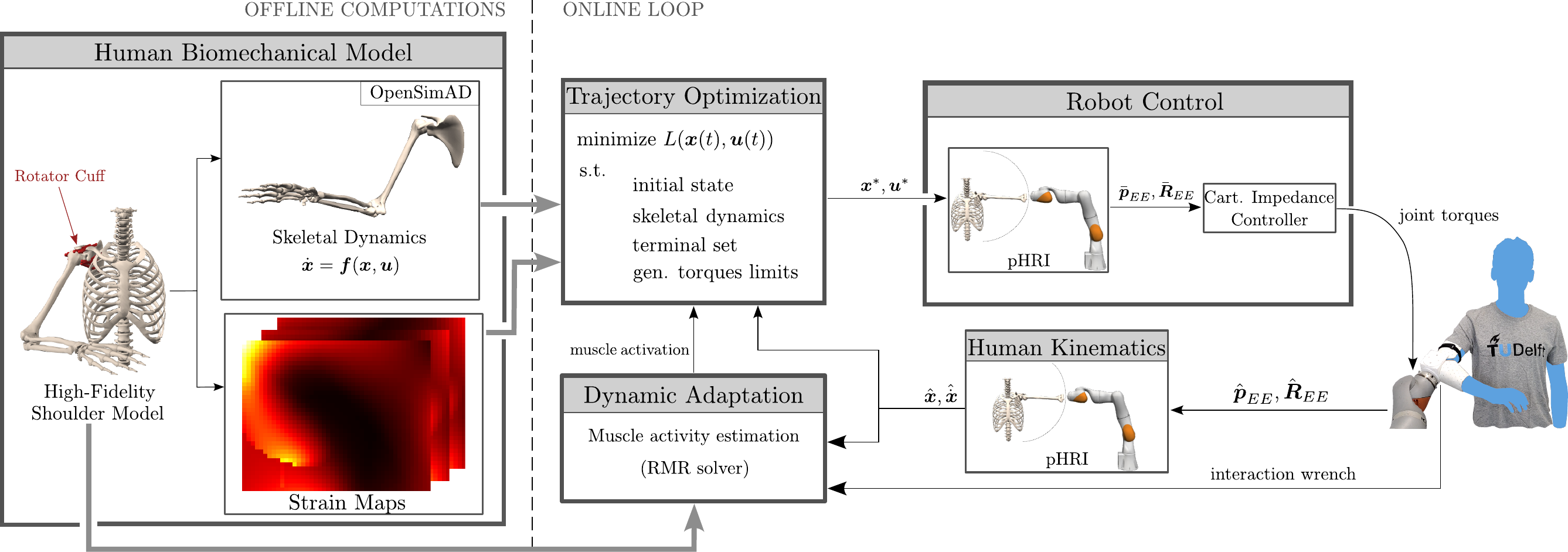}
    \caption{Methodological overview of BATON applied to physiotherapy for the rotator cuff muscle tendons of the human shoulder. The high-fidelity biomechanical model of the human shoulder (left), \addedText{employed fully to estimate human muscle activations,} is decoupled in skeletal dynamics and rotator cuff tendon loading information (strain maps) for efficient trajectory planning. An optimal control problem computes safe rehabilitation trajectories leveraging insights from the human model, and its output is transformed into equivalent references for the robotic controllers to track during physical human-robot interaction (pHRI). \addedText{The robot's feedback on the current human status is used to close the loop, allowing real-time estimation of the physiological activation for the muscles of interest.}}
    \label{fig:block_scheme}
    \vspace{-0.3cm}
\end{figure*}

\addedText{Our method, BATON (see Fig.~\ref{fig:block_scheme}, for overview) extracts insights from the musculoskeletal model for real-time use in optimization and control (Sec.~\ref{subsec:human_msk_model}). To exercise the range of motion of the human shoulder during therapy, the planner finds the minimum strain path by solving a trajectory optimization problem (Sec.~\ref{subsec:musculoskeletal_TO}). However, the physical interaction between the robot and the human during trajectory execution induces loads that the human responds to voluntarily and/or reflexively. Since human decisions, sensitivities, and responses to movement and other stimuli are inherently unpredictable, it is imperative that the human response in the control loop is based on actual measurements. 
Such human-in-the-loop actions and reactions change the level of muscle activity. Thus, muscle forces and corresponding tissue loads cannot be fully predicted, but need to be incorporated. To this end, we estimate these actions and reactions online using a state-of-the-art muscle redundancy solver (Sec.~\ref{subsec:rmr_solver}).} We employ a collaborative robotic arm \addedText{instrumented with a force-torque sensor} to estimate the current human pose \addedText{and loading} (Sec.~\ref{subsec:hum_state_estim}) and administer rehabilitation movements to the human subject. We embody the robot controller in a closed-loop with the subject (Sec.~\ref{subsec:robot_control_gravity_compensation}) to perform physical experiments (Sec.~\ref{subsec:experimental_design}) that evaluate our improvements over the state-of-the-art rotator cuff rehabilitation techniques.

\addedText{\section{Motion Planning on Strain Maps}} \label{sec:plan}
\addedText{Musculoskeletal insights are derived from a human biomechanical model, becoming the strain maps navigated by our online trajectory planner. 
We developed a rehabilitation path planner to traverse these maps to synthesize safe robotic-led human movements by formulating and efficiently solving an optimal control problem.}

To give a practical analogy of how we navigate safety during rehabilitation, let us consider the path planning of a ship in a bay where underwater reefs are present. To know which maneuvers allow the ship to navigate safely and efficiently toward a target position, we need both a dynamic model of the vessel and a map of the underwater reefs, providing non-obvious insights about the safest path. In our situation, the human shoulder dynamics play the role of the vessel, while the strain maps offer insights into what happens beneath the surface (in the tissue being rehabilitated). Safe navigation can only be obtained by considering where it is safe to travel, together with the maneuverability of the ship.

\subsection{Human Biomechanical Model}
\label{subsec:human_msk_model}


We employ a high-fidelity biomechanical model of the human shoulder~\cite{seth2019muscle}, developed in OpenSim~\cite{delp2007opensim, seth2018opensim}, to capture unpredictable volitional actions of the human and/or responses to motions and forces induced by robot interactions. The human model captures the shoulder's movement degrees of freedom (DoFs) and the muscles that generate movement. The model, therefore, can inform the robot's controller of the internal human mechanics during the physical human-robot interaction. In particular, we are interested in quantities related to injury risk of the rotator cuff muscles (i.e., supraspinatus, subscapularis, infraspinatus, and teres minor), which act as stabilizers for the glenohumeral (GH) joint in the shoulder and whose tendons are subject to tears due to overload, chronic weakness, and impingement. \addedText{With the model, we predict how movements and forces affect tendon strain in the rotator cuffs, which cannot be measured by surface EMG due to their position deep inside the body.}
Mechanical strain is associated with re-injury risk of rotator cuff tendons, as excessive stretching of a healing tendon could lead to tearing~\cite{chen2020retears}.

In the human body, the rotator cuff spans the GH joint, which permits mobility of the upper arm (humerus) with respect to the shoulder blade (scapula) through 3 DoFs. 
To improve efficiency without sacrificing fidelity, we reduced the original model to the 3 DoFs \addedText{and 22 muscles that actuate the GH joint only}, to isolate the effects on the rotator cuff muscles (see Fig.~\ref{fig:block_scheme}, left).
The arm dimensions and mass of our human subject were used to personalize the model's properties through OpenSim's scaling functionalities and tabulated anthropometric data~\cite{drillis1964body}. We define the state of the resulting human model as:
\addedText{
\begin{IEEEeqnarray}{c}
    \boldsymbol{\theta} = [PE, SE, AR]^{T} \\
    \boldsymbol{\alpha} = [\alpha_1, \alpha_2, ..., \alpha_{22}]^{T}, \ \alpha_i \in [0,1] \ \forall i\\
    \boldsymbol{x} = [\boldsymbol{\theta}^{T}, \dot{\boldsymbol{\theta}}^{T}, \boldsymbol{\alpha}^{T}]^{T} \label{math:full_human_state}
\end{IEEEeqnarray}
}
where $PE, SE$, and $AR$ are the 3 movement DoFs defined as the Y-X'-Y'' sequence of intrinsic Euler angles in the fixed shoulder reference frame ($PE$: plane of elevation, $SE$: shoulder elevation as rotation around -X, $AR$: axial rotation, visualized in Fig.~\ref{fig:coordinate_system} \addedText{and indicated collectively as $\boldsymbol{\theta} \in \mathbb{R}^{3}$), $\dot{\boldsymbol{\theta}} \in \mathbb{R}^{3}$ is the vector containing their derivatives with respect to time, and $\boldsymbol{\alpha}$ contains the muscle activation level for each of the muscles in the model. Note that, in the remainder of this paper, we will refer to $\boldsymbol{\mathrm{q}} = [\boldsymbol{\theta}^{T}, \dot{\boldsymbol{\theta}}^{T}]^{T}$ as the purely kinematic state of the model, while $\boldsymbol{x}$ is the complete musculoskeletal model state.}

The personalized shoulder model specifies how the shoulder moves under the influence of externally applied torques/forces, and how these affect the underlying musculoskeletal system. However, embedding the model directly to plan rehabilitation movements through the OpenSim API is not computationally tractable for real-time applications. To achieve real-time performances without sacrificing model accuracy, we decouple tissue mechanics that describe tendon strain with shoulder movement and muscle activation from skeletal dynamics that simulate how passive shoulder motion is generated as a consequence of externally applied forces. This decoupling enables us to formulate the problem of navigating shoulder motion over strain maps governed by skeletal dynamics and to solve planning efficiently.

\subsubsection{Navigable strain maps}
\addedText{A strain map describes how strains in individual tendons, or collectively  (e.g. the peak), change with respect to the model state (motion and activation). As the complete human model state $\boldsymbol{x}$ spans a high-dimensional space, we focus on the pose (target position from the therapist) and overall level of activation as the primary dimensions of a spatial strain map. We compute the strain map querying the} musculoskeletal model to extract quantitative information on the mechanical strain $\sigma_i$ of every i\textsuperscript{th} tendon in the rotator cuff. Biologically, tendon strain depends on the model's state $\boldsymbol{x}$ as:
\begin{equation}
    \sigma_i = \frac{l_{i}(\boldsymbol{x}) - l_{i,0}}{l_{i,0}} \cdot 100\%.
    \label{math:tendon_strain}
\end{equation}
where $l_{i}(\boldsymbol{x})$ is the current length of the tendon and $l_{0,i}$ its slack length at rest.
We fully describe the change in strain $\sigma_i$ based on the human model's state via pre-computed ``strain maps"~\cite{prendergast2021biomechanics, balvert2022enabling, beck2023real}, for every tendon of interest.
In this way, we created high-dimensional strain maps for every individual tendon, which allowed us to target only one tendon at a time or, alternatively, to monitor the strain on the rotator cuff as a whole by constructing aggregate maps that consider the highest strain across all the tendons~\cite{prendergast2021biomechanics, beck2023real}. To simplify the understanding of strain level for non-expert users of our system (e.g., physiotherapists and patients), we chart $\sigma_i$ on 2D grids where the strain percentage is visualized as a function of $PE$ and $SE$ angles, while both $AR$ and $\alpha_i$ are fixed. To consider different values of $AR$ and $\alpha_i$, other (pre-computed) 2D maps can be traversed. 


To traverse over strain maps efficiently, we interpolated our discrete strain maps using 2D Gaussian radial basis functions.
Through least-squares fitting, we obtain the parameter vector $\boldsymbol{\pi} \in \mathbb{R}^{N_p \times N_G}$ that describes the combination of the $N_G$ Gaussian functions best approximating every 2D strain map, in terms of the $N_p$ parameters of every Gaussian.

\subsubsection{Controlling skeletal dynamics}
After computing navigable differentiable maps of rotator cuff tendon strain, we account for the human movement dynamics when the human shoulder is being manipulated by the robot. \addedText{Human dynamics should be used to inform the trajectory planner (Sec.~\ref{subsec:musculoskeletal_TO}) about the relationship between forces/torques that the robot applies to the human, and the resulting human accelerations. However, the OpenSim API provides this relationship only in a non-differentiable form, which led previous research~\cite{supernumerary2021} to employ less efficient gradient-free optimization methods to optimize human-robot interaction.
To efficiently represent human skeletal dynamics for control purposes, we employed OpenSimAD~\cite{falisse2019algorithmic} as a tool that generates differentiable functions from OpenSim musculoskeletal models. OpenSimAD extends OpenSim by recording the call sequence used to compute specific outcome variables within an OpenSim model, enabling algorithmic differentiation through frameworks like CasADi~\cite{Andersson2019casadi}.} In particular, we customized the OpenSimAD pipeline to obtain a differentiable representation of our human model in terms of the ordinary differential equations capturing the dynamics of its multibody skeletal system, given a set of generalized forces $\boldsymbol{u} \in \mathbb{R}^N$ and the model's kinematic state $\boldsymbol{\mathrm{q}} \in \mathbb{R}^{N_q}$:
\begin{equation}
    \dot{\boldsymbol{\mathrm{q}}} = f(\boldsymbol{\mathrm{q}}, \boldsymbol{u}),
    \label{math:sys_dyn}
\end{equation}

where $N = 3$ and $N_q = 2 N$ in our case.
Importantly, this procedure guarantees that the model's joint definitions will be preserved, together with personalized dimensions, mass, and inertia for every existing bone in the original model.

\subsection{Biomechanics-aware trajectory optimization}
\label{subsec:musculoskeletal_TO}
\
We combine information from the differentiable strain maps and human skeletal dynamics to formulate a biomechanics-aware optimal control problem (OCP), with the goal of synthesizing safe human motion towards a target shoulder position. The solution to the OCP is the optimal trajectory $\boldsymbol{\mathrm{q}}^*$ and corresponding generalized forces $\boldsymbol{u}^*$ driving the human model toward the goal pose. \addedText{Note that we optimize generalized forces rather than end-effector wrenches to keep our trajectory optimization agnostic to how rehabilitation trajectories are delivered to the patient. The actual reference for the robot controller is obtained from} the optimal sequence of kinematic states and controls $\{\boldsymbol{\mathrm{q}}^*, \boldsymbol{u}^*\}$ in Sec.~\ref{subsec:robot_control_gravity_compensation}.

We opted to plan trajectories that are fully contained in a single 2D strain map, so that the output of our optimization would be intuitive for non-expert users and easy to visualize in two dimensions (e.g., on a computer screen). We selected plane of elevation ($PE$) and shoulder elevation ($SE$) to be our controlled DoFs, while current values of axial rotation ($AR$) and muscle activation $\boldsymbol{\alpha}$ are input for trajectory planning to select the correct strain map.

We define below the cost function and constraints of the OCP that allow us to set the optimality criterion with respect to which $\{\boldsymbol{\mathrm{q}}^*, \boldsymbol{u}^*\}$ are found, and present how the OCP is solved in practice.
\subsubsection{Cost Function}
we design our cost function to capture important requirements for a safe rehabilitation trajectory: the movement should minimize the strain on selected tendons, produce low accelerations on the human body to reduce interaction forces and discomfort, and evolve towards the target final position. These elements are mirrored into the definition of our cost $L(\boldsymbol{x}, \boldsymbol{u})$, formalized as a weighted sum of the following terms:
\begin{itemize}
    \item $L_{\sigma} = \sigma(\boldsymbol{x})$, accounting for the instantaneous (non-negative) strain that the selected tendon group undergoes;
    \item $L_{\text{acc}} = \dot{\boldsymbol{\mathrm{q}}}^{T} \boldsymbol{Q_1} \dot{\boldsymbol{\mathrm{q}}}$, accounting for the derivatives of our relevant state variables during the motion. The positive semidefinite matrix $\boldsymbol{Q_1} \in \mathbb{R}^{6 \times 6}$ selects only joint accelerations when planning on the current strain map;
    \item $L_{\text{$\tau$}} = \frac{1}{d_0^2}(\boldsymbol{\mathrm{q}}-\boldsymbol{\mathrm{q}}_{\text{$\tau$}})^{T} \boldsymbol{Q_2} (\boldsymbol{\mathrm{q}}-\boldsymbol{\mathrm{q}}_{\text{$\tau$}})$, accounting for the distance to the target human model pose $\boldsymbol{\mathrm{q}}_{\tau}$, where $d_0$ represents the distance to the target pose at the beginning of the trajectory, to normalize the contribution of this term. Again, $\boldsymbol{Q_2} \in \mathbb{R}^{6 \times 6}$ is a positive semidefinite matrix selecting only the relevant human DoFs. 
\end{itemize}
Overall, the cost to be minimized is:
\begin{equation}
    L(\boldsymbol{x}, \boldsymbol{u}) = w_{\sigma} \sigma(\boldsymbol{x}) + w_{\text{acc}} \dot{\boldsymbol{\mathrm{q}}}^{T} \boldsymbol{Q_1} \dot{\boldsymbol{\mathrm{q}}} + w_{\tau} \frac{1}{d_0^2} \boldsymbol{\Delta}_{\mathrm{q}}^{T} \boldsymbol{Q_2} \boldsymbol{\Delta}_{\mathrm{q}}
\end{equation}
where $w_{\sigma}$, $w_{\text{acc}}$ and $w_{\tau}$ are scalars weighting the contribution of the various terms, and we indicated $(\boldsymbol{\mathrm{q}}-\boldsymbol{\mathrm{q}}_{\text{$\tau$}}) = \boldsymbol{\Delta}_{\mathrm{q}}$ for brevity.

\subsubsection{Constraints}
\label{subsubsec:ocp_const}
We apply constraints to ensure that the OCP solution can be executed safely on the human subject. Their definition and role is detailed below.
\begin{itemize}
    \item[-] \textit{Initial conditions}: every new trajectory $\{\boldsymbol{\mathrm{q}}^*, \boldsymbol{u}^*\}$ should start from the current kinematic state of the human model $\hat{\boldsymbol{\mathrm{q}}}_{\text{curr}}$, estimated through the robot's encoders (see Sec.~\ref{subsec:hum_state_estim}). This value is used as the initial condition for the next instance of the OCP enforcing $\boldsymbol{\mathrm{q}}(t=0) = \hat{\boldsymbol{\mathrm{q}}}_{\text{curr}}$, where the initial time is set to be 0 without loss of generality.
    \item[-] \textit{Torque limits}: to ensure that following the optimal trajectory does not require excessive torque to be exerted along the human DoF, we limit them within heuristic bounds by requiring $\boldsymbol{u}_{\text{min}} \leq \boldsymbol{u}(t) \leq \boldsymbol{u}_{\text{max}}$, where $\boldsymbol{u}_{\text{min}}, \boldsymbol{u}_{\text{max}} \in \mathbb{R}^3$. Bounds can be adjusted to enforce different torque limits on the various DoF accounting, for example, for the fact that torques along SE should counteract gravity.
    \item[-] \textit{Terminal condition}: to impose an acceptable final state for the human model, we prescribe that $(\boldsymbol{\mathrm{q}}_{t=T_f} - \boldsymbol{\mathrm{q}}_{\tau}) \odot (\boldsymbol{\mathrm{q}}_{t=T_f} - \boldsymbol{\mathrm{q}}_{\tau}) \leq \boldsymbol{\epsilon}$, where $T_f$ represents the length of the planning horizon, and $\odot$ the Hadamard product. If reaching the target final state is possible in the current OCP instance (e.g. if $T_f$ is long enough), the parameter $\boldsymbol{\epsilon} \in \mathbb{R}^6$ is defined for every element of $\boldsymbol{\mathrm{q}}$, meaning both final pose and velocity of the human model can be specified fully. Otherwise, if the target pose specified by $\boldsymbol{\mathrm{q}}_{\tau}$ is too far from $\hat{\boldsymbol{\mathrm{q}}}_{\text{curr}}$, only the final velocities are limited. This choice guarantees low velocities of the human arm at the end of the planning horizon and ensures recursive feasibility through a safe terminal set $\mathbb{S}_T$ when the optimization is performed iteratively over a receding time horizon.
\end{itemize}

\subsubsection{Optimal control problem (OCP) transcription}
Overall, the optimal control problem that we aim to solve reads as follows:
\begin{equation} 
\label{math:ocp}
\begin{aligned}
& \underset{\boldsymbol{\mathrm{q}}(\cdot), \boldsymbol{u}(\cdot)}{\operatorname{min}} \ \ \int_0^{T_f} L\big(\boldsymbol{x}(t),\boldsymbol{u}(t)\big)dt\\
& \text{subject to:}\\ 
\end{aligned}
\end{equation}
\vspace{-0.5cm}
\begin{IEEEeqnarray*}{lc}
    \quad \boldsymbol{\mathrm{q}}(0) - \hat{\boldsymbol{\mathrm{q}}}_{\text{curr}} = \boldsymbol{0} & \ \text{initial state} \\
    \quad \boldsymbol{u}_{\text{min}} \leq \boldsymbol{u}(t) \leq \boldsymbol{u}_{\text{max}} & \ \text{torque limits} \\ 
    \quad \boldsymbol{\mathrm{q}}_{t=T_f} \in \mathbb{S}_T & \ \text{terminal set}
\end{IEEEeqnarray*}

To solve the OCP, we cast it into discrete time, transcribing~\eqref{math:ocp} into an equivalent Non-Linear Programming problem (NLP) that can be solved by structure-exploiting solvers. We used orthogonal collocation techniques~\cite{garg2017overview}, hence approximating the state trajectories with suitable d\textsuperscript{th}-order polynomial splines. The overall optimization horizon $T_f$ is broken down into $N$ intervals of equal length, and inside the generic interval $[t_k, t_{k+1}]$ we select $d=3$ Legendre-Gauss collocation points at which the dynamics as in~\eqref{math:sys_dyn} are enforced. This step results in additional continuity and collocation constraints that need to be considered in the NLP formulation. Resulting optimal trajectories $\{\boldsymbol{\mathrm{q}}^*, \boldsymbol{u}^* \}$ were upsampled for smoother execution.

We implemented the resulting NLP in Python and solved it with IPOPT~\cite{wachter2006implementation} leveraging MA27~\cite{ma27_hsl} as a linear solver, while derivatives were provided by CasADi~\cite{Andersson2019casadi} and OpenSimAD. To clarify the computational advantage of using the differentiable skeletal dynamics, we also embedded numerically evaluated gradients with perturbation calls to the original OpenSim model directly in the collocation constraints of the NLP, through CasADi. We ran simulations for both cases to compare computational requirements.

\addedText{\section{Dynamic Adaptation to Human Actions}}

\addedText{The trajectory on strain maps generated by the planner (Sect.~\ref{subsec:musculoskeletal_TO}) provides a reference for the robot controller. Nevertheless, unpredictable volitional actions of the patient and/or reflexive reactions to physical human-robot interaction can induce large interaction forces and corresponding muscle activations that lead to potentially large changes in the strain map. In this section, we describe how we exploit the robot's sensory system to make real-time updates on estimates of patient pose and muscle activity. These estimates are used to update the strain maps, which the planner will use to re-plan the robot trajectories in an online manner to avoid dangerous high-strain areas and abide by given optimization objectives.}

\addedText{\subsection{Online estimation of muscle activity}}
\label{subsec:rmr_solver}
\addedText{Estimating the activity of deep rotator cuff muscles is not feasible using traditional surface electromyography (EMG) measurements. To overcome this limitation, we employed a model-based estimation approach to infer individual muscle activations within our biomechanical model during robot-assisted rehabilitation. Specifically, we used the Rapid Muscle Redundancy (RMR) solver~\cite{belli2023does, beck2023real}, which yields physiologically plausible muscle activations in over-actuated systems — i.e., systems with more muscles than degrees of freedom — while accounting for the stabilizing role of the rotator cuff at the glenohumeral joint~\cite{belli2023does}.
As the human musculoskeletal model is inherently redundant, the solver leverages nonlinear optimization to estimate muscle activations consistent with observed kinematics and forces. The RMR solver minimizes a muscle effort cost function subject to physiological constraints to generate realistic recruitment patterns.}

\addedText{The original open-source implementation of RMR was designed for offline use. In contrast, our application requires real-time estimation, with the model state updated continuously based on measured joint kinematics (Section~\ref{subsec:hum_state_estim}) and external human-robot interaction wrenches (Section~\ref{subsec:experimental_design}). To this end, we restructured the solver into a class-based module integrated with the Robot Operating System (ROS), enabling online estimation of muscle activations $\hat{\boldsymbol{\alpha}}$ in a dedicated sub-process. These real-time estimates were used to select the tendon strain map that informed the trajectory optimization described in Sec.~\ref{subsec:musculoskeletal_TO}.}

\addedText{The code implementation is available in the project repository at \url{https://anonymous.4open.science/r/baton-robotic-rehab-2545}.} \\

\subsection{Human kinematics estimation} \label{subsec:hum_state_estim}


\begin{figure}[t!]
    \centering
    \includegraphics[width=\columnwidth]{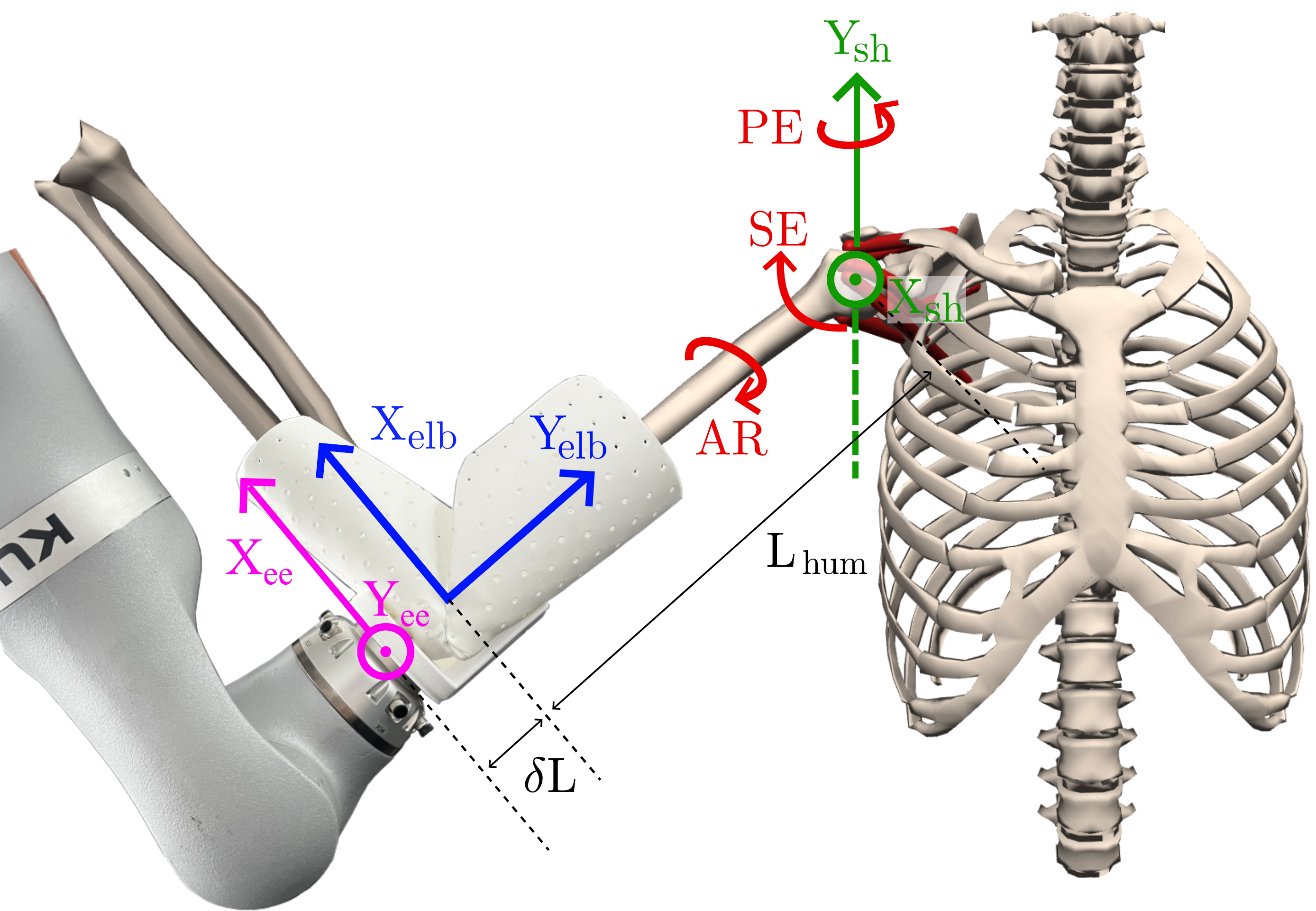}
    \caption{\addedText{The human and robot coordinate systems and their correspondence.} The shoulder frame (green) originates in the center of the glenohumeral joint. The glenohumeral joint DoFs (PE, SE, and AR) are shown in red. The elbow frame (blue) originates in the center of the elbow, and a fixed transformation relates it to the robot's end-effector frame (pink).}
    \label{fig:coordinate_system}
    \vspace{-0.4cm}
\end{figure}

We estimated the human model's kinematic state $\hat{\boldsymbol{\mathrm{q}}}$ through the position, orientation, and twist of the robot's end-effector (EE), which we mapped to the shoulder state $\boldsymbol{x}$ according to the fixed rigid connection between the EE and elbow locked at $90^\circ$(see Fig.~\ref{fig:coordinate_system}). An estimation of the full human body pose is out of the scope of our current work, so we assume for simplicity that the center of the glenohumeral joint is stationary during the experiments and that the orientation of the human torso is fixed. Under these conditions, monitored during the experiments, it is possible to reconstruct the shoulder kinematic state through the robot's encoders. 

We observe that the orientation of the human elbow expressed in the reference frame of the human shoulder is\footnote{we denote as ${}^{I}\boldsymbol{p}_{k} = [{}^{I}p_k^x, {}^{I}p_k^y, {}^{I}p_k^z]^T \in \mathbb{R}^3$ the position of point $k$ w.r.t. frame ``I", and ${}^{\text{J}}\boldsymbol{R}_{\text{I}} \in \mathit{SO}(3)$ the rotation matrix to express it in frame ``J".} :
\begin{equation}
    {}^{\text{sh}}\boldsymbol{R}_{\text{elb}} = {}^{\text{sh}}\boldsymbol{R}_{\text{base}} {}^{\text{base}}\hat{\boldsymbol{R}}_{\text{EE}}(t) {}^{\text{EE}}\boldsymbol{R}_{\text{elb}},
\end{equation}
where \textit{sh} denotes the shoulder frame, \textit{base} the frame attached to the robot's base, and \textit{elb} the frame fixed at the human elbow (see Fig.~\ref{fig:coordinate_system}).
The first and last rotations are known and fixed, while ${}^{\text{base}}\hat{\boldsymbol{R}}_{\text{EE}}(t)$ can be estimated through the robot's forward kinematics. 
Moreover, following the definition of the glenohumeral DoFs in the human model~\cite{seth2019muscle}, we can write:
\begin{equation}
    {}^{\text{sh}}\boldsymbol{R}_{\text{elb}} = \boldsymbol{R}_Y(PE) \boldsymbol{R}_X(-SE) \boldsymbol{R}_Y(AR).
    \label{math:gh_rotation_matrix}
\end{equation}

We first obtain our estimates $\hat{PE}$ and $\hat{SE}$ leveraging the end-effector position ${}^{\text{base}}\hat{\boldsymbol{p}}_{EE}(t)$, and then estimate $\hat{AR}$ through~\eqref{math:gh_rotation_matrix}. Formally, we define a unit vector $\boldsymbol{\beta} \in \mathbb{R}^3$ expressing the direction of the humerus (upper arm) pointing from the center of the glenohumeral joint to the human elbow, and use its Cartesian components in the shoulder frame to estimate the human state:
\begin{IEEEeqnarray}{lc}
    \boldsymbol{\beta} = {}^{\text{sh}}\boldsymbol{R}_{\text{base}} \frac{{}^{\text{base}}\hat{\boldsymbol{p}}_{EE}(t) - {}^{\text{base}}\boldsymbol{p}_{\text{GH}}}{||{}^{\text{base}}\hat{\boldsymbol{p}}_{EE}(t) - {}^{\text{base}}\boldsymbol{p}_{\text{GH}}||_2^2} \nonumber \\
    \hat{PE} = \mathrm{atan2}(\beta^x, \beta^z) \\
    \hat{SE} = \mathrm{arccos}(\boldsymbol{\beta} \cdot [0 \ -1 \ 0]^T) \nonumber
\end{IEEEeqnarray}
where ${}^{\text{base}}\boldsymbol{p}_{\text{GH}}$ is the known position of the center of the glenohumeral joint. 
Velocities of the generalized DoFs in the human model are estimated through the body twist of the last link of the robot, obtained through the robot's Jacobian:
\begin{equation}
    {}^{\text{base}}\hat{\boldsymbol{\nu}}_{\text{EE}} = [{}^{\text{base}}\hat{\boldsymbol{v}}_{\text{EE}}^T \quad {}^{\text{base}}\hat{\boldsymbol{\omega}}_{\text{EE}}^T]^T \in \mathbb{R}^6.
\end{equation}

Since the robot's EE and the human elbow are rigidly attached to each other, we can transform ${}^{\text{base}}\hat{\boldsymbol{\nu}}_{\text{EE}}$ into the twist ${}^{\text{sh}}\hat{\boldsymbol{\nu}}_{\text{elb}}$ through the known transformation:
\begin{equation}
    {}^{\text{sh}}\hat{\boldsymbol{\nu}}_{\text{elb}} = \begin{bmatrix} \boldsymbol{R} & 0\\ 0 & \boldsymbol{R} \end{bmatrix} {}^{\text{base}}\hat{\boldsymbol{\nu}}_{\text{EE}}
\end{equation}

where $\boldsymbol{R} = {}^{\text{sh}}\boldsymbol{R}_{\text{elb}} {}^{\text{elb}}\boldsymbol{R}_{\text{EE}} {}^{\text{EE}}\boldsymbol{R}_{\text{base}}$.
Having retrieved $\hat{PE}$ and $\hat{SE}$, ${}^{\text{sh}}\hat{\boldsymbol{\nu}}_{\text{elb}}$ can be further transformed into ${}^{\text{sh'}}\boldsymbol{\nu}_{\text{elb}}$, denoting the elbow twist expressed in the intermediate rotated frame where $SE$ is defined (which we name \textit{sh'}), and into ${}^{\text{sh''}}\boldsymbol{\nu}_{\text{elb}}$, which denotes the elbow twist expressed in the frame where $AR$ is defined (which we name \textit{sh''}).
Overall, we can obtain the velocity estimates for the human shoulder model as follows:
\begin{IEEEeqnarray}{lc}
    \hat{\dot{PE}} =  {}^{\text{sh}}\omega_{\text{elb}}^y, \qquad \text{with} \quad {}^{\text{sh}}\boldsymbol{\omega}_{\text{elb}} = \frac{{}^{\text{sh}}\boldsymbol{p}_{\text{elb}} \times {}^{\text{sh}}\hat{\boldsymbol{v}}_{\text{elb}} }{L_{\text{tot}}^2} \nonumber \\
    \hat{\dot{SE}} =  -{}^{\text{sh'}}\omega_{\text{elb}}^x, \quad \text{with} \quad {}^{\text{sh'}}\boldsymbol{\omega}_{\text{elb}} = \frac{{}^{\text{sh'}}\boldsymbol{p}_{\text{elb}} \times {}^{\text{sh'}}\hat{\boldsymbol{v}}_{\text{elb}} }{L_{\text{tot}}^2}  \\
    \hat{\dot{AR}} = {}^{\text{sh''}}\hat{\omega}_{\text{elb}}^{y}    \nonumber
\end{IEEEeqnarray}
where $L_{\text{tot}} = L_{\text{hum}} + \delta L$ is the distance between the center of the glenohumeral joint and the robot's EE (see Fig.~\ref{fig:coordinate_system}), ${}^{\text{sh}}\boldsymbol{p}_{\text{elb}}$ is the position of the human elbow expressed in the shoulder frame, and ${}^{\text{sh'}}\boldsymbol{p}_{\text{elb}}$ is the equivalent quantity expressed in the \textit{sh'} frame.

\addedText{Finally, we differentiated numerically $\hat{\dot{PE}}$, $\hat{\dot{SE}}$, and $\hat{\dot{AR}}$ to obtain the corresponding acceleration vector $\boldsymbol{\ddot{\theta}}$. Exponential filtering was applied to the estimated values to reject high-frequency components incompatible with human movements in physiotherapy.}

\section{\addedText{Robot control and experimental design}}
\addedText{To test BATON as a novel approach to robotic-assisted rotator cuff rehabilitation and demonstrate its capabilities, we designed and performed various lab-based experiments. These proof-of-concept scenarios are essential for the validation of our approach as a whole, testing BATON against the state of the art and in situations where its dynamic adaptation becomes crucial for guaranteeing safe rehabilitation.}

\subsection{Robot control}
\label{subsec:robot_control_gravity_compensation}
In addition to human state estimation, our collaborative robotic arm administers therapeutic motion to the human subject. We map the optimal trajectory $\boldsymbol{\mathrm{q}}^*$ resulting from~\eqref{math:ocp} into the corresponding robot's end-effector reference pose:
\begin{IEEEeqnarray}{lc}
    {}^{\text{base}}\bar{\boldsymbol{R}}_{\text{EE}} = {}^{\text{base}}\boldsymbol{R}_{\text{sh}}  {}^{\text{sh}}\boldsymbol{R}_{\text{elb}}(\boldsymbol{\mathrm{q}}^*) {}^{\text{elb}}\boldsymbol{R}_{\text{EE}} \label{math:robot_reference_rot} \\
    {}^{\text{base}}\bar{\boldsymbol{p}}_{\text{EE}} = {}^{\text{base}}\boldsymbol{p}_{\text{GH}} + {}^{\text{base}}\bar{\boldsymbol{R}}_{\text{EE}} {}^{\text{EE}}\boldsymbol{p}_{\text{GH}} \label{math:robot_reference_cartesian}
\end{IEEEeqnarray}
Equations~\eqref{math:robot_reference_rot}-\eqref{math:robot_reference_cartesian} ensure that the robot's end-effector pose is consistent with the desired trajectory for the human. Given the optimal shoulder angles, the EE pose follows from the human elbow position and orientation, and is oriented such that the upper arm always points to the center of the human shoulder.

\addedText{To track target end-effector poses, we used an impedance controller that governs the interaction force/torque in Cartesian space as:
\begin{equation}
    \boldsymbol{f} = \boldsymbol{K} (\bar{\boldsymbol{X}}_{\text{EE}}-\hat{\boldsymbol{X}}_{\text{EE}}) - \boldsymbol{D} \hat{\dot{\boldsymbol{X}}}_{\text{EE}}, \label{eq:impedancecontrol}
\end{equation}
where $\boldsymbol{f}$ is the interaction force/torque between the robot and its patient, $\bar{\boldsymbol{X}}_{\text{EE}}\in \mathbb{R}^6$ is the commanded EE reference retrieved from ${}^{\text{base}}\bar{\boldsymbol{R}}_{\text{EE}}$ and ${}^{\text{base}}\bar{\boldsymbol{p}}_{\text{EE}}$, and $\hat{\boldsymbol{X}}_{\text{EE}}\in \mathbb{R}^6$ is the measured actual pose of the robot, respectively. $\boldsymbol{K}\in \mathbb{R}^{6 \times 6}$ and $\boldsymbol{D}\in \mathbb{R}^{6 \times 6}$ are the Cartesian stiffness and damping matrix, respectively. The impedance controller operates at $200$ Hz.}

\addedText{In addition to the Kuka's built-in gravity compensation, we include further compensation to support the mass of the subject's arm. Specifically, we use the optimal generalized torques from the OCP in~\eqref{math:ocp} to adjust the vertical reference for the robot's EE, increasing it proportionally to the torques required to produce the optimal movement. This adjustment, further detailed in the \hyperref[sec:appendix]{Appendix}, allows our system to keep the actual vertical position of the human arm close to the prescribed one, leveraging pure impedance control without the need to feed-forward assistive torques.}

\subsection{Experimental design for evaluation of BATON}
\label{subsec:experimental_design}

Our experimental setup consisted of a KUKA LBR iiwa 7 robotic arm \addedText{instrumented with a Bota SensONE force-torque (F/T) sensor (Bota Systems AG, Zurich, CH)} and a custom thermoplastic elbow brace for interfacing the human subject with the robot. The brace was molded to cradle the subject’s elbow (Fig. 3) and mounted directly onto the F/T sensor at the robot's end-effector, fixing the transformation between the robot EE frame and the human elbow frame. This configuration allowed the robot to impose motion on the human arm, support its weight, and simultaneously estimate the subject’s biomechanical state.

A healthy adult subject participated in our lab-based experiments, approved by 
the Human Research Ethics Committee at Delft University of Technology. 
The brace was strapped to the participant’s arm, and they were instructed to maintain a fixed torso posture relative to the robot base, satisfying the assumption of a stationary glenohumeral joint center required by our state estimation model.

To evaluate BATON and highlight the importance of modeling human arm dynamics in trajectory planning, we examined two rehabilitation scenarios: (1) the early stages of therapy, during which the robot provides full guidance while the subject remains passive; and (2) the later stages, where the subject progressively regains strength and actively participates in the motion. \addedText{The first scenario also served to benchmark BATON against the state-of-the-art (SoTA) method proposed by Prendergast et al.~\cite{prendergast2021biomechanics}, which generates strain-aware shoulder rehabilitation movements based on tendon behavior. Their method uses a modified A\textsuperscript{*} algorithm to traverse discrete passive strain maps while minimizing cumulative strain, returning a kinematic sequence of optimal joint angles $\boldsymbol{\mathrm{q}}^{\text{SoTA},*}$. We implemented this planner in our framework and compared its performance with BATON's in guiding a passive subject. The second scenario, involving an active subject, falls outside the applicability of the SoTA method, which assumes constant muscle activations and thus cannot accommodate unpredictable activation changes.}

\subsubsection{Passive human subject}

In this scenario, the subject remained relaxed, allowing the robot to lead the movement. Muscle activity was negligible, and tendon strain was primarily determined by the shoulder pose. To ensure uniformly low strain across the rotator cuff, we monitored a strain map that aggregated the maximum strain values over the supraspinatus, infraspinatus, subscapularis, and teres minor tendons. The strain value at a given model state $\boldsymbol{x}$ was defined as the maximum of these individual tendon strains.
To maintain consistency on the strain map, the axial rotation (AR) degree of freedom (DoF) was locked, so that the complete movement toward the target arm pose $\boldsymbol{q}_{\tau}$ evolved along a single map. \addedText{Negligible muscle activation and} constant axial rotation imply that trajectories can be optimized and then executed in an open loop. Multiple target poses $\mathbf{q}_{\tau}$ were defined and used as inputs to BATON to explore a broader range of motion at the glenohumeral joint. Each optimization instance used a time horizon of $T_f = 5$ seconds, discretized into $N = 50$ intervals. Cost function weights were selected based on extensive simulations to provide parameter tuning guidelines.

All candidate trajectories were first tested in the Gazebo simulator to verify the robot's ability to execute them accurately.
\addedText{We then compared BATON-generated trajectories to those from the SoTA method, each executed on the physical robot. The SoTA trajectory $\mathbf{q}_\text{SoTA,*}$ was upsampled to match the execution frequency of BATON's output and fed to the controller via Eqs. (14)–(15). Performance was evaluated in terms of strain minimization, trajectory smoothness, and human–robot contact forces. Both methods started from the same human shoulder pose $\boldsymbol{q}(0)$ and navigated between the same target poses.}

\subsubsection{Active human subject}
\addedText{Due to human volitional actions and/or reflexive reactions during therapy, the level of muscle activity and forces can change quickly and significantly, impacting tendon strain estimates. BATON needed to adapt its trajectory planning in real time to maintain safety under these dynamic conditions. We shortened the planning horizon to $T_f = 1$ second, with $N = 10$ intervals, allowing for continuous replanning using a receding horizon scheme.}

\addedText{We first used simulations to investigate the reactiveness of BATON to artificial muscle activation changes when treating a specific tendon. Without loss of generality, the infraspinatus was considered. The initial human pose was set at $SE_0=PE_0=60^\circ$, while the target position to $\bar{PE}=45^\circ$, $\bar{SE}=95^\circ$, requiring movement along multiple human DoFs. Simulation also allowed us to consider drastic variations on fabricated maps to trigger more extreme trajectory adaptations as a consequence, for example, of a strong reflex response. In this case, the previous initial and target positions were inverted, representing the previous movement that evolved in the opposite direction.}

\addedText{Then, we moved to a closed-loop execution of BATON on the real robot, employing realistic infraspinatus activity estimation $\hat{\alpha}_{\text{is}}$ through the RMR solver (Sec.~\ref{subsec:rmr_solver}).}

\addedText{In this scenario, we analyzed the effect that physical human-robot interaction and volitional human behavior have on online therapy adjustments. 
BATON guided our subject between the same initial and final poses ($SE_0 = PE_0 = 60^\circ$, $\bar{PE} = 45^\circ$, $\bar{SE} = 95^\circ$) under two conditions: (i) the subject complied with the robot, and (ii) the subject actively engaged the infraspinatus by actively externally rotating their arm. This rotation, while not substantially disrupting the robot-guided trajectory, was resisted by the Cartesian controller, which maintained the initial value of the $AR$ DoF. The resulting interaction force elicited infraspinatus activation, consistent with its primary anatomical function~\cite{terry2000functional}.
}

Model-based computations were run on a Dell Latitude 7420 laptop with an i7-1185G7 CPU, interfaced with a workstation running a Xeon W-2123 and executing the robot's impedance controller. This ensured low-level control at 200 Hz, decoupled from potential delays in high-level planning. \addedText{The Robot Operating System managed communications between the various software modules.}




\section{Results}

\addedText{We performed a thorough evaluation of the proposed methods. Regarding the planning aim (A1), we analyzed the effect of optimization parameters on the planned trajectory (Sec.~\ref{subsec:results_ocp_parameter_tuning}) and compared BATON to the SoTA method in terms of the quality of the executed paths (Sec.~\ref{subsec:results_quality_executed_paths_Astar_comparison}). Regarding adapting to changing muscle activity (A2), we demonstrated BATON's ability to re-plan online based on changing strain maps, first in simulation and then thanks to real-time muscle activity estimates (Sec.~\ref{subsec:results_online_adaptation_active_subject}). We concluded by conducting an analysis of real-time computational performance (Sec.~\ref{subsec:results_computational_performance}). 
}
\begin{figure}[t!]
    \centering
    \includegraphics[width=\columnwidth]{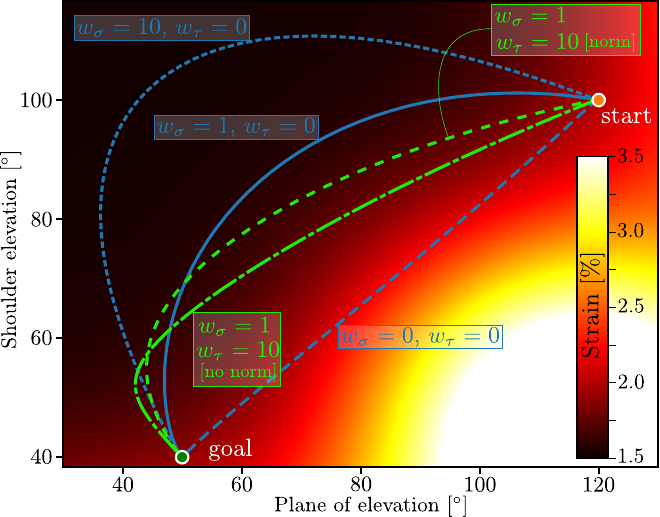}
    \caption{Effect of tuning the cost function parameters on the optimal trajectories for the human DoF. 
    In blue, the influence of $w_{\sigma}$ on the optimal path. In green, the effect of normalizing $L_{\tau}$ when weighting the distance to the target final pose. To guarantee slow motions, we kept $w_{\text{acc}}=10$.}
    \label{fig:exp1_simulation_tuning}
    \vspace{-0.4cm}
\end{figure}
\subsection{Effect of OCP parameters on trajectories (passive subject)}
\label{subsec:results_ocp_parameter_tuning}

\begin{figure*}[t!]
    \centering
    \includegraphics[width=0.9\textwidth]{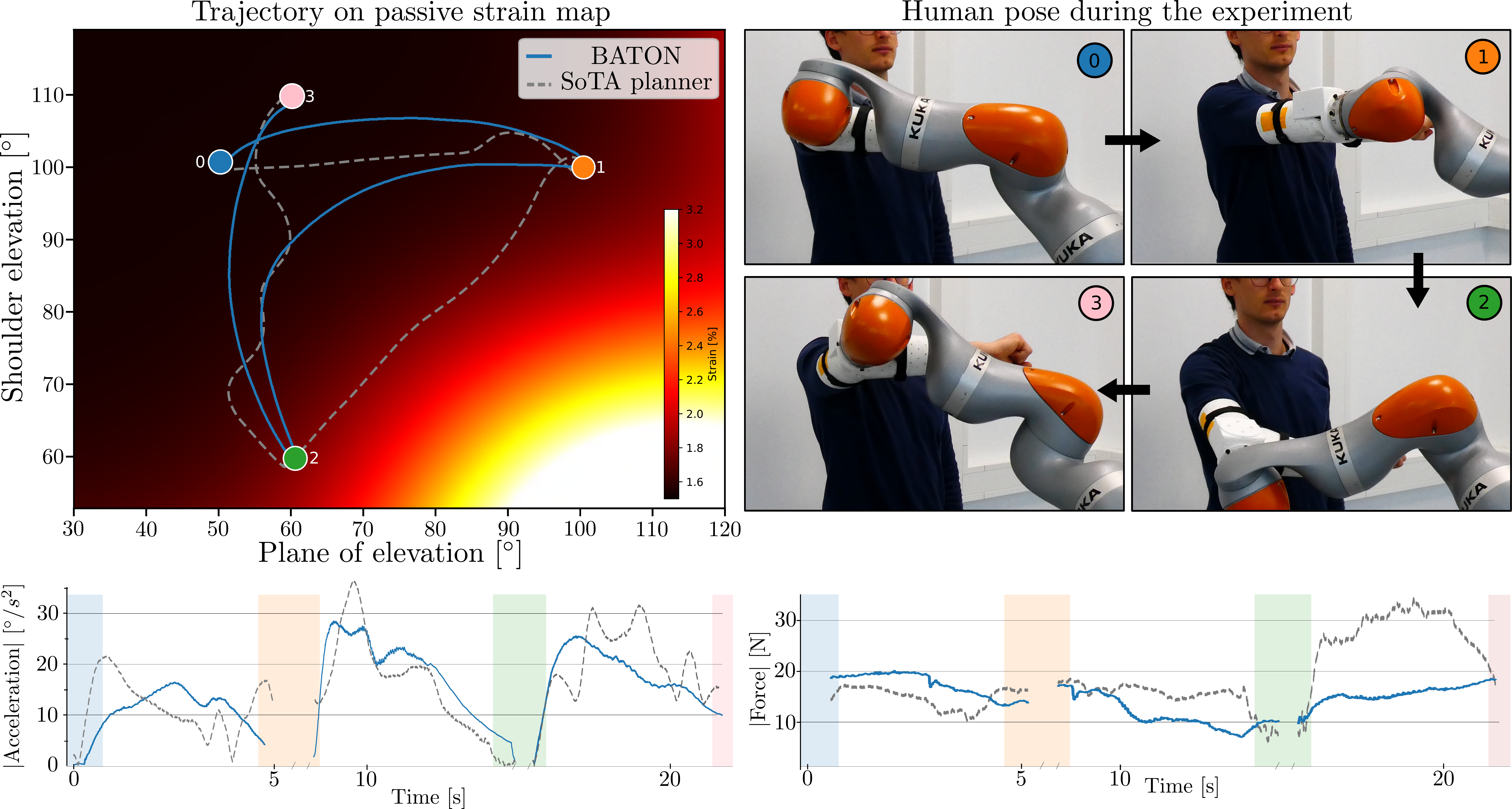}
    \caption{Top left: human trajectories produced by our approach for passive rehabilitation (solid blue), \addedText{visualized on the corresponding strain map together with the result of the planner in~\cite{prendergast2021biomechanics} (dashed grey)}. The two lines are the actual trajectory, in human coordinates, along which the robot moved our subject. The paths connect target human poses given as goals to 3 instances of our trajectory optimization problem minimizing the maximum strain that the rotator cuff tendons encounter, \addedText{and 3 instances of the planner in~\cite{prendergast2021biomechanics}. Top right: goal poses of the test subject during the experiment. The graphs at the bottom display the magnitude of the acceleration along the two trajectories (left) and the interaction force magnitude between the robot and subject during the motion (right). Note that the data corresponding to re-planning periods is not shown for conciseness.}}
    \label{fig:exp1_real_robot}
    \vspace{-0.3cm}
\end{figure*}

We first ran BATON in simulation to examine the effects of optimization parameters and to determine the best relative cost function weights for synthesizing rehabilitation movements for passively assisted patients. We explored various permutations of the cost function weights $w_{\sigma}$, $w_{\text{acc}}$, and $w_{\tau}$, modulating the importance that tendon strain, human accelerations, and movement towards the goal pose have in the overall cost function. Fig.~\ref{fig:exp1_simulation_tuning} shows some of the representative scenarios. We fixed $w_{\text{acc}} = 10$ and varied the other two parameters to achieve rather extreme cases. If minimizing the strain is ignored ($w_{\sigma}=0$), the planner would choose the shortest path (dashed blue line). As higher weights $w_{\sigma}$ are used, the trajectory makes a wider excursion in the low-strain regions (solid and dotted blue lines). While this guarantees safer therapy, we observed that increasing this term gradually produces wider trajectories while the quantitative decrease in strain becomes negligible. In other words, wider excursions do minimize the cost function value, but they lead to insignificant reductions in strain. Given this trade-off, we fixed $w_{\sigma}=1$ for later experiments.

With a passive subject, the strain map was fixed and did not vary with time. Thus, we could directly optimize the entire trajectory and execute it in an open-loop manner. Here, both final human pose and velocities are constrained (see Sec.~\ref{subsubsec:ocp_const}), so the target pose is reached even when $w_{\tau} = 0$. This terminal constraint effectively simplifies the tuning of our cost function to only two parameters in this case. By selecting $w_{\text{acc}}=10$, $w_{\sigma}=1$ the resulting trajectory traverses safe low-strain regions ($\sigma \leq 2 \%)$. These weights were used for the robot experiments in Sec.~\ref{subsec:results_quality_executed_paths_Astar_comparison} below.

However, enforcing a terminal constraint on the final human pose is undesirable with shorter planning horizons, which are needed to react to changes in the strain map as muscle activations vary (active subject scenario, Sec.~\ref{subsec:results_online_adaptation_active_subject}). A trajectory deemed safe on the initial map may later intersect high-strain regions as the map is updated. This motivates setting $w_{\tau} \neq 0$, and investigating the effect of normalizing the distance to the target $L_{\tau}$ on the optimal trajectory (Fig.~\ref{fig:exp1_simulation_tuning}, green lines). When $L_{\tau}$ was normalized by the initial distance from $\boldsymbol{\mathrm{q}}_{\tau}$, lower-strain trajectories were produced, prioritizing strain minimization over approaching the goal earlier. This normalization ensures that the goal pose provides a movement direction, and prevents this information from vanishing when the goal is closer or dominating the cost when $\boldsymbol{\mathrm{q}}_{\tau}$ is farther.

\subsection{Quality of executed path w.r.t. SoTA}
\label{subsec:results_quality_executed_paths_Astar_comparison}
\addedText{Resulting trajectories from BATON were compared to the ones obtained with the SoTA planner from the literature~\cite{prendergast2021biomechanics}. Figure~\ref{fig:exp1_real_robot} shows the experimental setup and the target poses on the strain map, together with the optimal trajectories that were delivered to our test subject by the robot in both cases. For both executions,} the robot started at the shoulder pose marked by the blue dot in the low-strain zone. First, it moved the patient to the pose marked by the orange dot and then toward the state indicated by the green dot to increase the level of strain in a controlled manner and augment the subject's range of motion. Finally, the robot moved the subject back into the low-strain zone denoted by the pink dot.

BATON's resulting trajectory avoided higher strain regions (Fig.~\ref{fig:exp1_real_robot}, top left), navigating the rather large range of motion with lower strain overall. \addedText{On the other hand, the SoTA method produced trajectories that tend to cross zones of higher strain, and lead to oscillations in the subject's movement (especially evident towards the last pose).
The bottom row of Fig.~\ref{fig:exp1_real_robot} (left) shows the smoothness of the resulting movements in terms of human acceleration magnitude, with the SOTA approach producing consistently higher accelerations with respect to BATON. The evolution of the interaction forces between our subject and the robot is also shown for both approaches (bottom right), with BATON's trajectory requiring lower interaction forces, especially during the last part of the movement.}

\subsection{Online adaptations (active subject)}
\label{subsec:results_online_adaptation_active_subject}

\begin{figure}[t!]
    \centering
    \includegraphics[width=\columnwidth]{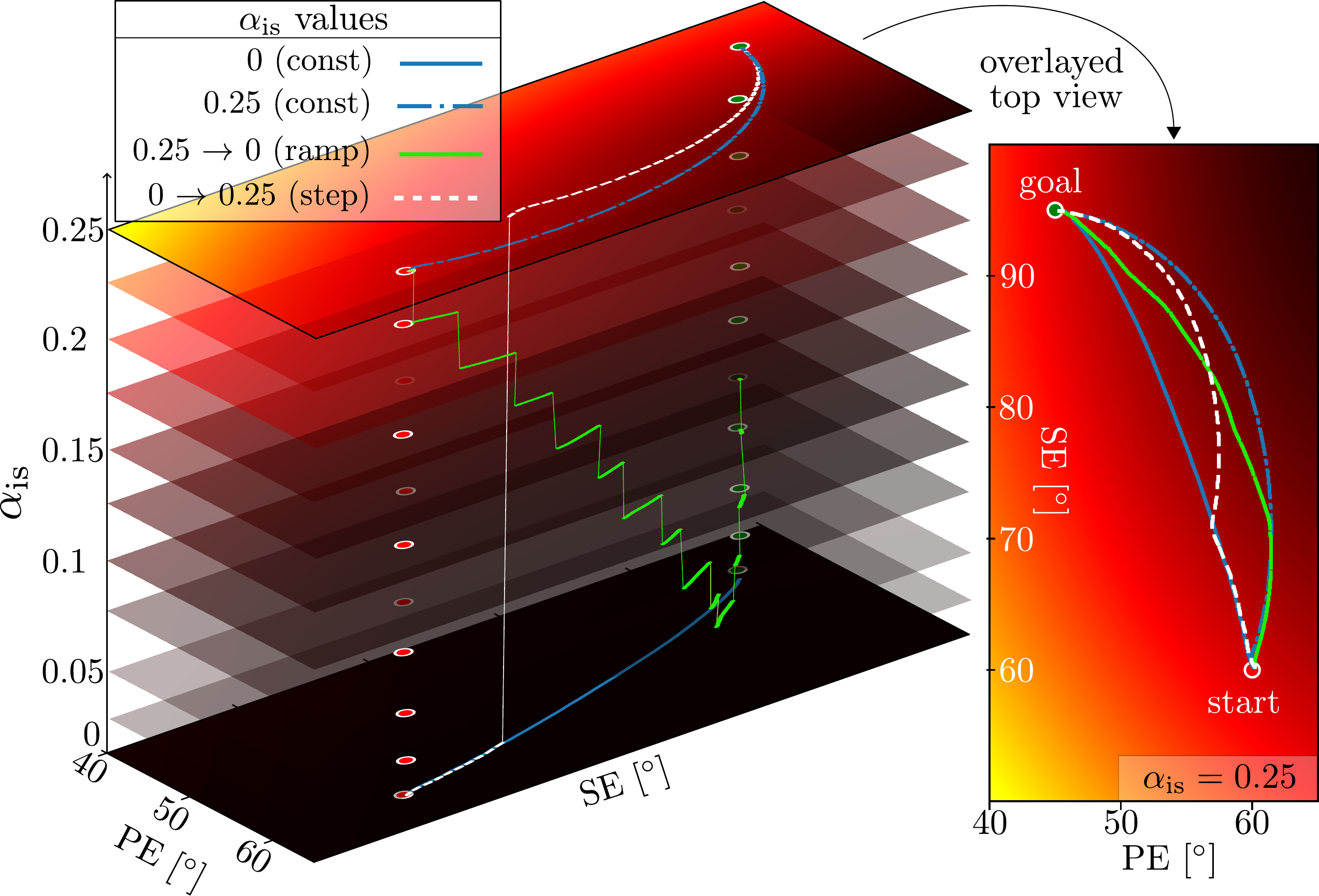}
    \caption{The simulated effects of infraspinatus muscle activation ($\alpha_{\text{is}}$) on the optimal trajectory with the same target pose, when $\alpha_{\text{is}}$ follows different activation profiles (green: decreasing ramp, white: step) or remains constant (blue). Large activation steps are used for ease of visualization, and all of the resulting trajectories are projected on the strain map corresponding to the highest $\alpha_{\text{is}}$ for comparison to the right. Our musculoskeletal trajectory optimization is capable of accounting for sudden variations in the strain level, as a consequence, e.g., of varying muscle activation.}
    \label{fig:exp2_simulation_activation_modifies trajectory}
\end{figure}

\subsubsection{Simulations}
First, we investigated the responsiveness and adaptability of our control scheme when the strain maps change as a consequence of artificial changes in infraspinatus activations.
Owing to the shorter time horizon over which we ran our trajectory optimization, BATON was able to achieve a trajectory re-planning with an update rate of roughly 10 Hz. This permits us to generate new trajectories that can account for strain map variations, allowing us to control human movement in a closed-loop manner.

The effect of two different muscle activation changes is visible in Fig.~\ref{fig:exp2_simulation_activation_modifies trajectory}: a sudden step increase in activation (white trajectory) and a gradual decrease in activation (green trajectory). These two cases are shown alongside the two boundary cases where activation is kept constantly high or low (resulting in the blue trajectories in the figure). We visualize all of the trajectories jointly on a single strain map, representing the strain map defined by the highest activation, set at $\alpha_{\text{is}} = 0.25$ here (Fig.~\ref{fig:exp2_simulation_activation_modifies trajectory}, right). Re-projecting all of the trajectories on the same plane highlights differences due to the gradual and sudden variations of the activation level, bounded by the two cases in which the activation was kept constant.

\begin{figure}[t!]
    \centering
    \includegraphics[width=\columnwidth]{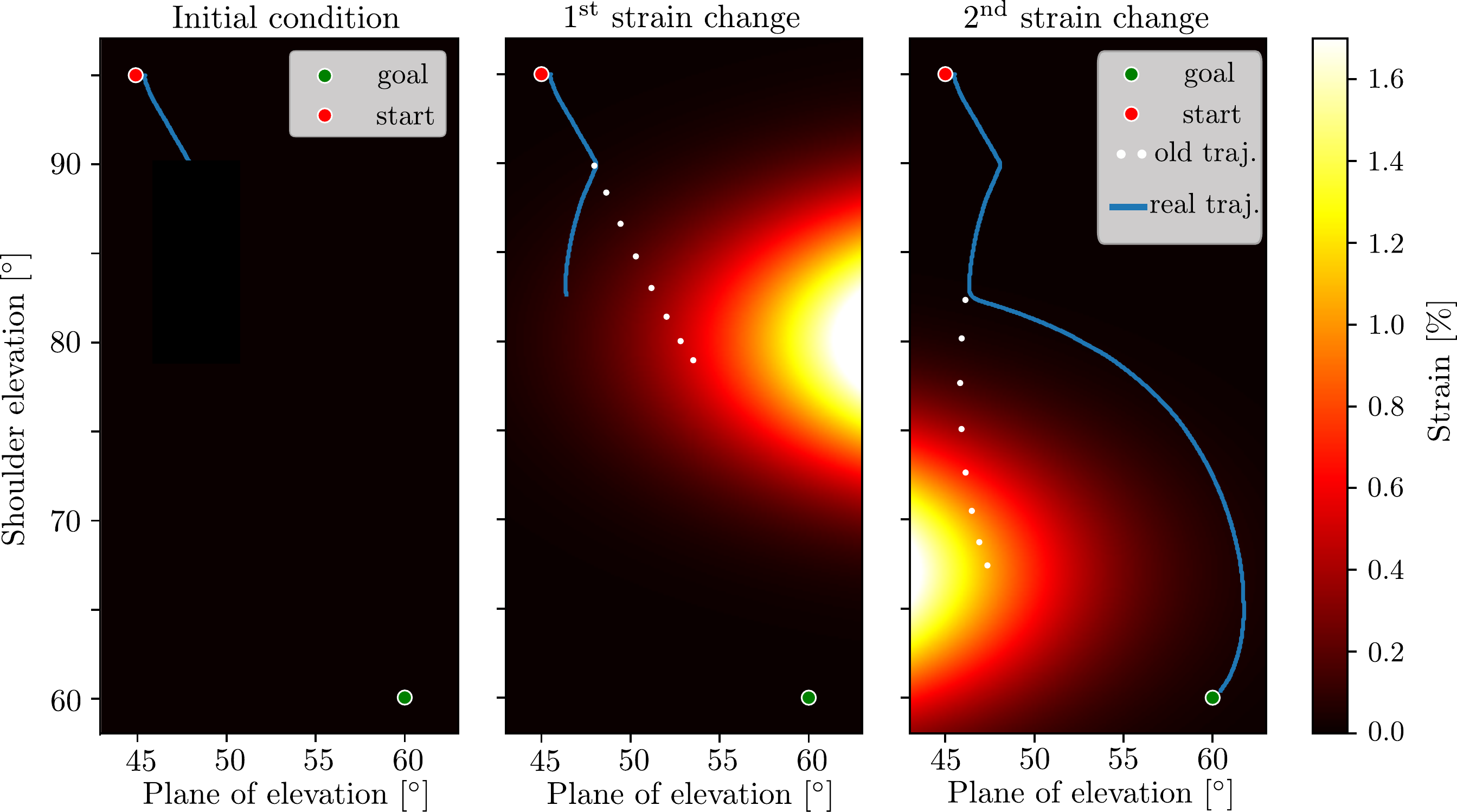}
    \caption{Real-time trajectory optimization allows to account for sudden variations in the strain level. We show here how the optimal trajectory changes as a consequence of rapid variations of (fabricated) strain maps. The trajectory optimization starts on a completely safe strain map, generating a motion directly toward the goal (left). Sudden variations of the maps cause the reference to avoid higher-strain zones (center and right), generating a trajectory that deviates from the shortest path adapting to the new scenarios, and prioritizing safety.}
    \label{fig:exp2_simulation_sudden_variations}
\end{figure}

To understand the effects of more sudden and large changes in activation, for example, due to a strong reflex response, we tested cases in which BATON is presented with larger, more localized, and sudden variations in the strain maps (see Fig.~\ref{fig:exp2_simulation_sudden_variations}). First, the strain map on which to plan has low strains, so the optimal trajectory points directly to the goal (Fig.~\ref{fig:exp2_simulation_sudden_variations}, left). We then introduced a sudden large local variation in the strain map topography, so that higher strain appears along the direction of movement. BATON reacted to the variation by re-planning a new trajectory that keeps evolving towards the lower-strain region, deviating from the older plan proposed at the previous time step (white dots in the figure). Similarly, we introduced a second sudden large local variation in the strain map, observing a new trajectory adaptation that allows the simulated movement to remain safe.

\subsubsection{Real system}
\begin{figure*}[t!]
    \centering
    \includegraphics[width=1.5\columnwidth]{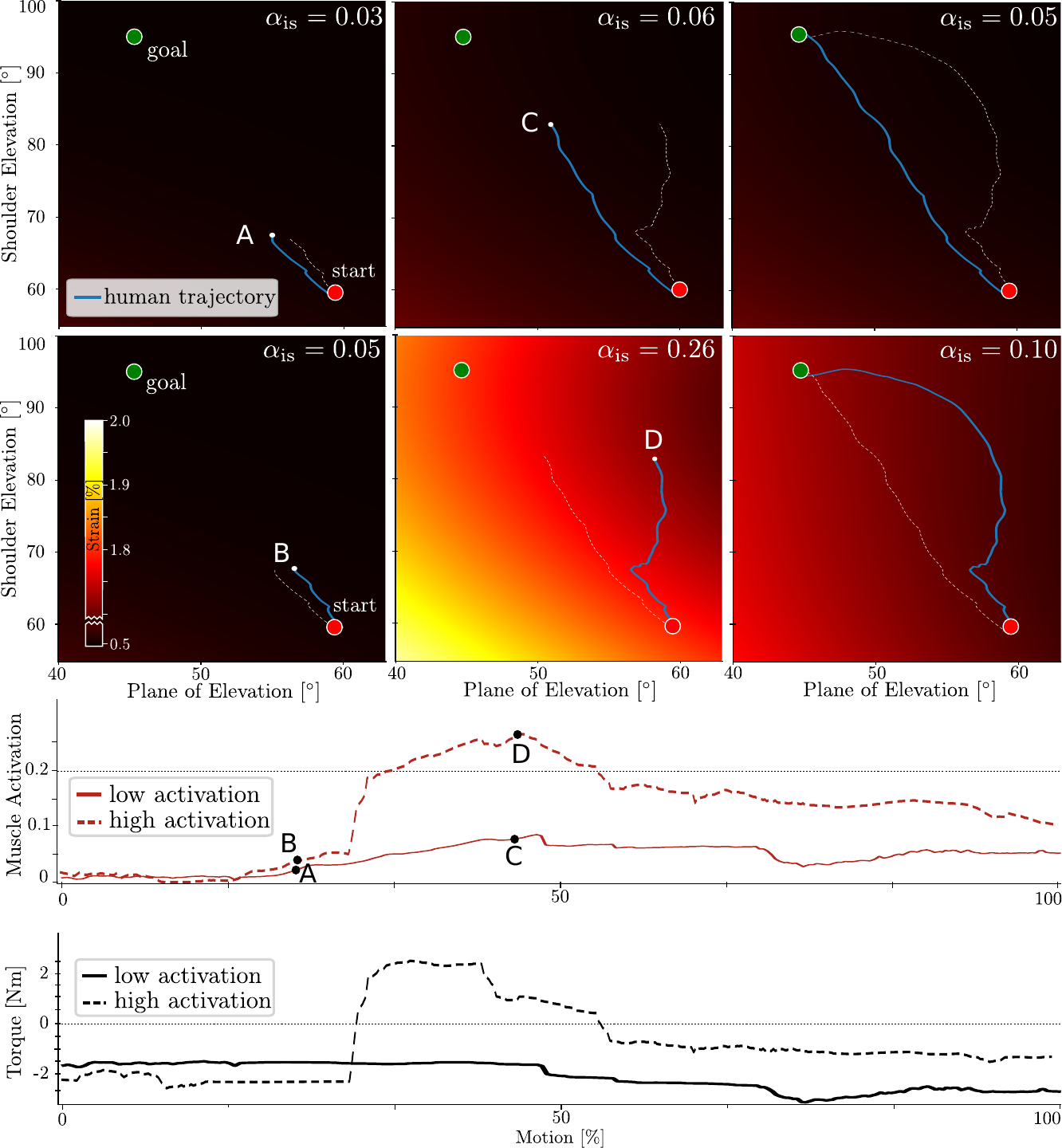}
    \caption{\addedText{Different \textit{infraspinatus} activation $\alpha_{\text{is}}$ causes different strain map topography for the corresponding tendon in our physical human-robot interaction experiments. Feeding human pose and interaction torque to BATON, it adapts to changes in estimated muscle activity. Here, we reveal the progression of rehabilitation trajectories along varying strain maps for our subject complying with the robot guidance (top row) and voluntarily activating their infraspinatus (second row). The solid blue lines represent the trajectory executed for each case, while dashed white lines allow us to compare them to the other case. Bottom rows: evolution of estimated infraspinatus activation $\hat{\alpha}_{\text{is}}$, and interaction torque modulated by the subject during the motion for both cases.}}
    \label{fig:exp2_real_robot}
    \vspace{-0.3cm}
\end{figure*}

Having explored and compared the real-time adaptation of the reference optimal rehabilitation trajectories in simulation, we proceeded to demonstrate this in a real robotic-assisted experiment. \addedText{Here, online estimates for all of the muscles in the human model were achieved through the RMR solver (Sec.~\ref{subsec:rmr_solver}) at 20 Hz. Based on the estimated infraspinatus activation $\hat{\alpha}_{\text{is}}$, the strain map considered by trajectory optimization was dynamically updated at runtime. The subject followed the robot-led movement twice, between the same starting and goal shoulder poses. In the first condition, they remained passive as the robot moved their arm, and the strain map topography was mostly static during execution (Fig.~\ref{fig:exp2_real_robot}, first row). In the second condition instead, they were instructed to perform an external rotation of their arm, resisted by the robot, to elicit infraspinatus activation (Fig.~\ref{fig:exp2_real_robot}, second row). Selected snapshots of the two movements, visualized on the time-varying strain maps, reveal BATON's adaptations to muscle activation changes. In particular, the two executions lead to a difference of about $10^\circ$ in human poses, as a result of adaptations of the human movements to traverse lower tendon strain areas defined by the time-dependent muscle activity values (Fig.~\ref{fig:exp2_real_robot}, third row) resulting from increased human-robot contact wrenches (Fig.~\ref{fig:exp2_real_robot}, bottom row).
}

\subsection{Computational performances}
\label{subsec:results_computational_performance}
Finally, we used simulations to compare the computational efficiency of our choice to capture the human skeletal dynamics with OpenSimAD, against numerically querying the original OpenSim model during the optimization (Sec.~\ref{subsec:human_msk_model}). This was done for an active subject as well as for a passive subject, modulating the time horizon for the trajectory optimization instances~\eqref{math:ocp}. The average computation times for solving~\eqref{math:ocp} when the system dynamics are enforced at the collocation points through the two different methods indicate a difference of roughly two orders of magnitude in favor of BATON (Tab.~\ref{tab:computation_time_comparison}). Real-time trajectory optimization was possible only with OpenSimAD, granting re-planning capabilities at a frequency of roughly $10$~Hz with appropriate selection of the planning horizon and its discretization. BATON achieved the same computational performances during the robot experiments.

\begin{table}[ht]
    \centering
    \caption{Computation time to solve a single OCP instance with different time horizons: A) passive subject, B) active subject}
    \vspace{-2mm}
    \begin{tabular}{c|c|c}
        Scenario & OpenSim+CasADi & OpenSimAD (BATON) \\ \hline
        A ($T_f=5$ s, $N=50$) & 227.0 s & 1.9 s \\
        B ($T_f=1$ s, $N=10$) & 8.50 s  & 0.12 s
    \end{tabular}
    \label{tab:computation_time_comparison}
\end{table}


\section{Discussion}

\subsection{\addedText{Motion Planning on strain maps}}
BATON is a new approach that enables robot-assisted physiotherapy to both automate and personalize rehabilitation by navigating underlying human biomechanical outcomes during therapy delivery. \addedText{Achieving our first research aim (A1),} BATON directly embeds human biomechanics into a robot controller through trajectory optimization and \addedText{adapts to muscle activation changes as they occur.} Accounting for high-fidelity musculoskeletal outcomes is crucial to deliver effective robotic rehabilitation. We developed and applied BATON to address the complexities and challenges of traditional physical therapy for rotator cuff injuries, where insights into patient-level tendon strains are paramount. Knowledge of the patient's tissue loads and range of motion is essential for optimizing safe movements that avoid high-strain configurations associated with risks of re-injury. These biomechanical insights allow us to plan and execute robotic rehabilitation trajectories, but can also be harnessed by physical therapists to guide the efficacy and safety of applied movements.


In both passive and active scenarios, we found BATON to operate several orders of magnitude faster than using traditional musculoskeletal modeling and simulation techniques (Tab.~\ref{tab:computation_time_comparison}). As a result, real-time re-planning of reference trajectories for the rehabilitation robot was attained at an average update rate of just below 10 Hz.

\subsection{\addedText{Dynamic adaptation to the human actions}}
\addedText{An integral part of the safety and potential autonomy of BATON is the feedback of the human response into the planning and control loop. We demonstrate the estimation and adaptation to changes in muscle activity in both simulation (Figs~\ref{fig:exp2_simulation_activation_modifies trajectory}-\ref{fig:exp2_simulation_sudden_variations}) and activity inducing movements in a test subject (Fig.~\ref{fig:exp2_real_robot}). The online model-based estimates of muscle activity account for changes in the patient's unpredictable volitional actions and/or reflexive responses due to physical human-robot interaction, achieving our second aim (A2). 
As a consequence, our system can inform the planner about underlying muscle activity, e.g., rotator cuff muscles, which cannot be measured by surface EMG. Although the proposed approach does not require surface EMG, such data could be used, if available, to augment the solver in estimating muscle activity.}

\addedText{We analyzed our algorithm over multiple edge cases in simulation (Fig.~\ref{fig:exp2_simulation_activation_modifies trajectory} and Fig.~\ref{fig:exp2_simulation_sudden_variations}), to verify the capability of our online trajectory optimization to cope with extreme strain map variations (potentially due to muscle activation changes). Then, we leveraged a model-based muscle estimation algorithm to close the loop between sensing, online trajectory optimization, and reference tracking on the real robot (Fig.~\ref{fig:exp2_real_robot}). The use of the RMR solver enabled our system to estimate physiological muscle recruitment without the use of instrumenting our subject with any kind of sensor (e.g., surface EMG ones), decreasing the time required for setup and enabling simultaneous monitoring of virtually any muscle in the human body (even deep stabilizers like the rotator cuff). Considering the infraspinatus muscle as a target tissue for our real-robot experiments, we achieved online therapy adaptations to its activation changes during physical human-robot interaction (Sec.~\ref{subsec:results_online_adaptation_active_subject}). Muscle activation was monitored both when the subject complied with the robot motion commands, and when they voluntarily elicited muscle activation. In the first case, rather small variations occurred, and they were likely due to the mechanical role of the infraspinatus as a glenohumeral stabilizer during humeral elevation. In the second case, voluntary muscle activation was estimated, in a scenario similar to real therapy when a patient's neuromuscular reflexes might be triggered, or the human might interact in unexpected ways with the robot.}

\subsection{\addedText{Evaluation insights of BATON performance}}
Further, we analyzed the performances of BATON through simulations and physical experiments \addedText{and achieved the third aim of this study (A3)}. The optimized trajectories can be tuned to reach a trade-off between the minimization of tendon strain and the directness of the path (Fig.~\ref{fig:exp1_simulation_tuning}). \addedText{BATON's optimized trajectories guaranteed lower instantaneous and cumulative tendon strain than the SoTA planner~\cite{prendergast2021biomechanics}, which instead traversed higher strain regions (Fig.~\ref{fig:exp1_real_robot}). Moreover, BATON produced smoother trajectories, while the SoTA planner delivered trajectories presenting higher oscillations, potentially in contrast with natural human movement preferences~\cite{flash1985coordination}. These oscillations are intrinsic to the assumptions of the SoTA kinematic planner, as it operates over a discrete version of the strain maps, resulting in various non-differentiable points along the commanded reference.
Beyond resulting in higher shoulder accelerations, the SoTA planner also applied higher peak forces to the human arm in our experiments (Fig.~\ref{fig:exp1_real_robot}, bottom row), a result that can be attributable to ignoring human dynamics during planning.}

The different modalities (i.e., an active and a passive subject) explored in this study can provide solutions for different stages of the therapy. After surgery or in early treatment after an injury, the patient is generally restricted in their mobility, and movements are carefully applied by the therapist who supports the arm and shoulder. We were able to control robot-led movements over long-horizon trajectories to increase the range of motion of the shoulder while minimizing tendon strain in the rotator cuff to avoid re-injury (Fig.~\ref{fig:exp1_real_robot}). As treatment progresses and as the patient regains strength and mobility, patients become and are encouraged to be more active during the exercises, in which case navigation must be more responsive (Figs.~\ref{fig:exp2_simulation_activation_modifies trajectory}-\ref{fig:exp2_simulation_sudden_variations}-\ref{fig:exp2_real_robot}).

In conclusion, our results demonstrate that embedding a high-fidelity representation of the human musculoskeletal system in the controller improves physical human-robot interaction in the context of automated yet responsive rehabilitation. In this study, we focused on rotator cuff strain monitoring and minimization, but other quantities could also be included (such as joint reaction forces or muscle activations themselves). Overall, the use of high-fidelity biomechanical models unlocks the design of robotic controllers that can monitor musculoskeletal quantities that even experienced therapists lack access to, potentially enhancing the utility of these systems in clinical rehabilitation. The next step is to evaluate the system on target users and patient groups.

\subsection{\addedText{Limitations and future work}}
We demonstrated and evaluated key aspects of BATON that provide confidence in its application for physiotherapy in the future. Nonetheless, we acknowledge that there are several limitations that require further investigation.
First, we focused on the mobility of the glenohumeral joint alone, and planned human motions along two of its DoFs ($PE$ and $SE$). In our setting, the motion along the third DoF ($AR$) of the human glenohumeral joint was not optimized, and the current value of $AR$ determined the strain map on which the optimization took place. This simplification allowed us to validate the core concept, but spatial (3D) rehabilitation trajectories should be investigated for use in therapeutic applications. For example, the strain of the rotator cuff tendon depends on the relative position between the upper arm and the shoulder blade, and other movement DoFs, such as the scapula mobility or the human torso orientation, will also influence tissue loading in the shoulder.

Second, the human pose estimation that we employed did not include the complete human state and may rely on simplifying assumptions that are too restrictive for a wide range of physical therapy applications. While an accurate estimation of the subject's body motion was beyond the scope of this work, better methods to monitor human posture and movement could be integrated. Viable options could rely on motion capture techniques, which offer high precision but are difficult to deploy, while video camera systems or IMUs might be more practical but also less reliable. Exoskeletons could also provide better human kinematics with respect to the collaborative robotics arm employed in this study. In the future, we plan to explore how to extend BATON to also include additional DoFs, such as those related to the scapula, with an approach to measure the state of the scapula itself.
However, apart from resulting in more complex computations, a challenging aspect of treating additional DoFs is delivering visual information that a patient/therapist can easily interpret.

\addedText{A third limitation lies in the strain value delivered by the shoulder model we selected. While the model has been validated in previous studies for what concerns the accuracy of shoulder kinematics~\cite{seth2016biomechanical} and muscle activity~\cite{seth2019muscle, belli2023does}, a thorough validation of the estimated tendon behavior across the model's range of motion is necessary before we can apply robotic physical therapy for rehabilitation from rotator cuff tendon tears on a patient population.
}

\addedText{Lastly, we did not predict human behavior at planning time, opting instead for online adaptation to volitional actions and reflexive reactions. While BATON could be extended to include predictive models of human motion or neuromuscular dynamics(e.g.,~\cite{sitole2023predictingjointmechanics, mugge2010rigorous}), they would provide only uncertain predictions that deviate from actual human behavior. The fundamental challenge of reliably predicting human decisions and behaviors remains. As such, we designed BATON to use actual behavior, necessary to ensure safety and effective performance, prioritizing real-time responsiveness to changes in human muscle activations estimated from measured movement and interaction forces. Future advances in biomechanical estimation could further enhance BATON’s performance.
}

\section{Conclusion}
We presented BATON, a novel approach to physical human-robot interaction and its application for shoulder rotator cuff physiotherapy. Through biomechanics-aware trajectory optimization, BATON generates and executes rehabilitation trajectories applied to a human subject by a collaborative robot, leveraging a personalized musculoskeletal model. Mechanical tendon strain was monitored and minimized both during the early, passive stages of rehabilitation, \addedText{and as muscle activation emerges when the subject responds, either actively or reactively, during therapy.} 

Unlike prior work in robotic-assisted physiotherapy, which does not directly monitor internal biomechanical metrics due to modeling and computational complexities, we have taken the first step to unlock the power of high-fidelity musculoskeletal models for real-time biomechanics-aware robotic controllers. \addedText{Moreover, integral to BATON is an online alternative to sensor-based muscle activity estimations, such as surface EMG, which cannot measure underlying muscles like the rotator cuff.} 

While our primary focus is robotic rehabilitation, the ability of automated systems to estimate \addedText{and respond to the changing internal states of the human body} resonates far beyond physical therapy. BATON represents a foundational shift toward the next generation of human-centered assistive and augmentative robotics. We pass the ``baton" to researchers and engineers in ergonomics, exoskeleton control, movement disorders, and beyond, where awareness of human biomechanics and subject/patient response is needed for robots to transform outcomes.

\addedText{\section*{Appendix}}
\label{sec:appendix}

\addedText{
Experimentally, commanding the Cartesian end-effector pose through~\eqref{math:robot_reference_cartesian} produces non-negligible tracking errors along the vertical axis if the human subject cannot support the weight of their own arm. Since full-arm support is typical of early rotator cuff rehabilitation~\cite{lastayo1998continuous}, we implemented a gravity-compensation strategy modifying~\eqref{math:robot_reference_cartesian} to include the generalized torques $\boldsymbol{u}^*$ into ${}^{\text{base}}\bar{\boldsymbol{p}}_{\text{EE}}$. Let us define the end-effector Cartesian position as ${}^{\text{base}}\bar{\boldsymbol{p}}_{\text{EE}} = {}^{\text{base}}\bar{\boldsymbol{p}}_{\text{GH}} + {}^{\text{base}}\bar{\boldsymbol{R}}_{\text{EE}}{}^{\text{base}}\bar{\boldsymbol{p}}_{\text{GH}} = [p_x \ p_y \ p_z ]^{T}$, similar to~\eqref{math:robot_reference_cartesian}. From our biomechanics-aware trajectory optimization (Sec.~\ref{subsec:musculoskeletal_TO}) we obtain the human generalized torques needed to track such a reference at every time instant. Thus, to deliver the required model-based torque along the $SE$ DoF to the human subject, we added to the vertical EE reference a displacement $\delta_z$:
\begin{equation}
     \delta_z = \frac{u^{*,\text{SE}}}{K_z L_{\text{hum}} \mathrm{sin}(\hat{\text{SE}})}
     \label{math:model_based_gravity_compensation}
\end{equation}
resulting in ${}^{\text{base}}\bar{\boldsymbol{p}}_{\text{EE}} = [p_x, \ p_y, \ p_z + \delta_z]^{T}$
}

\addedText{
This approach is personalizable to each subject through their specific biomechanics and appears more robust to model inaccuracies than purely torque-based compensation methods (i.e.~\cite{manzano2023model}). We show experimentally that this adjustment achieves the desired vertical EE setpoint, fully supporting the human arm without resorting to impractically high stiffness in the controller.}

\begin{figure}[t!]
    \centering
    \includegraphics[width=\columnwidth]{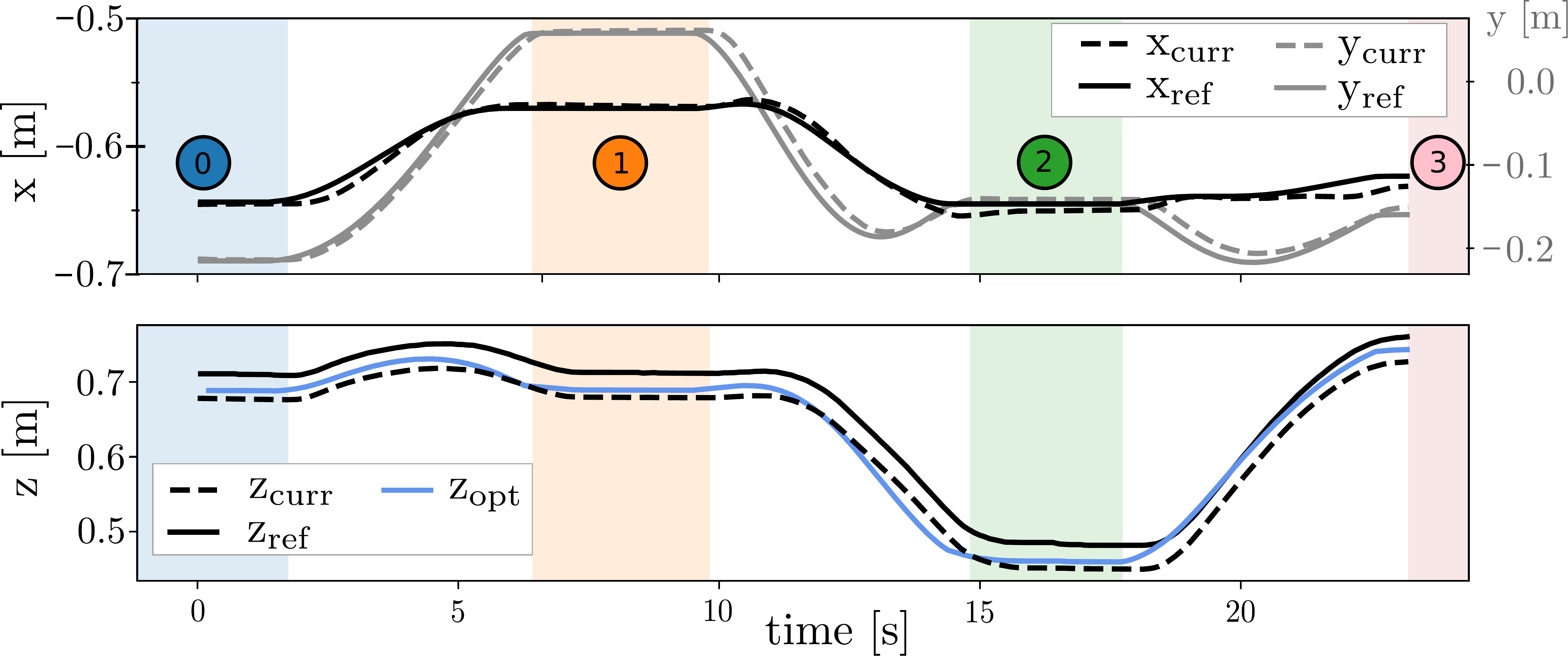}
    \caption{Tracking performances of BATON's optimal trajectory during passive therapy, expressed into the Cartesian coordinates of the robotic end-effector ${}^{\text{base}}\boldsymbol{p}_{\text{EE}}$. The bottom graph highlights the effect of using the personalized musculoskeletal model to inform the gravity compensation of the human arm: the offset reference given to the controller (solid black) makes it so that the current position (dashed) better tracks the optimal one (solid blue).}
    \label{fig:exp1_EE_robot}
    \vspace{-0.3cm}
\end{figure}

\addedText{Fig.~\ref{fig:exp1_EE_robot} shows the Cartesian reference positions for the robot EE together with the actual ones across the duration of the experiment. In particular, the bottom row of the figure presents the tracking performance along the robot's vertical axis. It can be noted that the vertical reference $z_{\text{}ref} = p_z + \delta_z$, accounting for the optimized offset in~\eqref{math:model_based_gravity_compensation}, is not reached. However, this allows the Cartesian impedance controller to provide the correct support, such that the optimal end-effector height $z_{\text{opt}} = p_z $ is tracked.
}

\bibliographystyle{IEEEtran}
\bibliography{bibliography}

\end{document}